%% file: acl_latex.tex
\documentclass[11pt]{article}

\usepackage[final]{acl}

\usepackage{times}
\usepackage{latexsym}
\usepackage{mdframed}

\UseRawInputEncoding
\usepackage[utf8]{inputenc}

\usepackage[T1]{fontenc}

\usepackage[utf8]{inputenc}

\usepackage{microtype}

\usepackage{inconsolata}

\usepackage{graphicx}
\usepackage{multirow}
\usepackage{booktabs}
\usepackage{amsmath}
\usepackage{amsfonts}
\usepackage{booktabs}
\usepackage[table]{xcolor}
\usepackage{colortbl}
\usepackage{xcolor}
\usepackage{array}
\usepackage{makecell}
\usepackage{enumitem}

\definecolor{bluedark}{RGB}{20, 50, 100}
\definecolor{bluelight}{HTML}{E8EEF7}
\definecolor{reddark}{RGB}{130, 30, 40}
\definecolor{redlight}{HTML}{F7E8E8}

\usepackage{listings}
\usepackage{tcolorbox}
\tcbuselibrary{listings, breakable}

\definecolor{promptbg}{RGB}{252, 250, 245}
\definecolor{promptframe}{RGB}{168, 155, 130}
\definecolor{titlebg}{RGB}{92, 118, 95}

\lstnewenvironment{promptbox}[1]{%
    \lstset{
        basicstyle=\ttfamily\fontsize{6.5}{7.8}\selectfont,
        breaklines=true,
        columns=fullflexible,
        keepspaces=true,
        frame=single,
        backgroundcolor=\color{promptbg},
        rulecolor=\color{promptframe},
        title={\colorbox{titlebg}{\color{white}\small\bfseries\strut\ #1\ }},
        aboveskip=10pt,
        belowskip=10pt,
        xleftmargin=4pt,
        xrightmargin=4pt,
        framesep=4pt,
    }%
}{}

\title{LexAgentHallu: A Hierarchical Benchmark for Profiling Hallucinations in Legal Agents}

\author{
\textbf{Yujin Zhou}$^{1*}$, \textbf{Mingxuan Zheng}$^{1*}$, \textbf{Chuxue Cao}$^{1*}$, \textbf{Yidan Huang}$^1$ \\
\textbf{Jiale Chen}$^1$, \textbf{Yike Guo}$^1$ , \textbf{Sirui Han}$^{1\dag}$\\
$^1$Hong Kong University of Science and Technology \\ 
\texttt{yzhouha@connect.ust.hk}
}

\newcommand{\ourbench}{LexAgentHallu}

\begin{document}
\maketitle
\begin{abstract}

As large language models are increasingly deployed as tool-augmented legal agents, they introduce agentic hallucinations where tool-call and reasoning errors cascade into fabricated holdings and miscited authority. However, existing legal benchmarks evaluate only single-turn QA with outcome-level metrics, while agentic hallucination benchmarks lack legal-specific diagnostic capability. Neither can determine to what extent and how a legal agent hallucinates along its trajectory. To address these limitations, we introduce \textbf{LexAgentHallu}, a legal agentic hallucination benchmark designed to evaluate to what extent and how legal agents fail along multi-step trajectories. Built through a four-stage expert-in-the-loop pipeline, LexAgentHallu contains 3{,}414 instances across 17 legal categories and 6 task types. Each instance is annotated under a dual-layer hallucination taxonomy of 7 mid-level categories and 27 fine-grained subclasses, covering both substantive errors and agent-procedural failures. We further design fine-grained metrics that quantify to what extent each failure occurs and identify how it occurs along an agent's execution path. Our evaluation across 18 proprietary and open-source agents uncovers a Right-Answer-Wrong-Reason effect and reveals that hallucination subclasses cluster rather than scatter, forming distinct agentic framework, legal task, and category profiles. These findings, invisible to outcome-level evaluation, validate the diagnostic power of LexAgentHallu for evaluating agentic hallucination in law. \footnote{Code will be available at \url{https://github.com/TOM-ZHOUch/LexAgentHallu}.}

\end{abstract}

\section{Introduction}

\input{sections/introduction}

\section{Related Work}

\input{sections/related_work}

\section{Benchmark}
\label{sec:benchmark}

\input{sections/benchmark}

\section{Experiments}
\label{sec:experiments}

\input{sections/experiments}

\section{Conclusion}

\input{sections/conclusion}

\section*{Limitations}
We acknowledge several limitations in the present work.
First, \textbf{LexAgentHallu} is grounded in the Chinese legal system: all queries, gold trajectories, and rubric checklists are written in Chinese and anchored to PRC statutes, judicial interpretations, and procedural rules. Because legal practice varies significantly across jurisdictions, and agentic legal tasks in common-law or mixed systems may surface hallucination patterns absent from our taxonomy—for example, errors in analogical reasoning from precedent, distinguishing binding from persuasive authority under stare decisis, or drafting jury instructions. This reflects a deliberate trade-off: fine-grained step-level attribution requires close alignment with jurisdiction-specific authority structures, which limits cross-jurisdictional generalizability. Future work will extend the taxonomy and benchmark to other legal systems through collaboration with legal-AI researchers across jurisdictions.

Second, as LLM capabilities continue to evolve, newer models may exhibit different performance and hallucination patterns on \textbf{LexAgentHallu} from those reported in this paper. To support continuous evaluation, we plan to launch an online platform that tracks the latest state-of-the-art legal agents and regularly updates per-subclass frequencies, co-occurrence patterns, and RAWR profiles, providing the community with up-to-date diagnostic results.

Finally, our benchmark experiments are limited to single-model legal agents. We make this choice to keep the evaluation setting controlled and to attribute hallucinations more clearly to individual reasoning trajectories, rather than to multi-agent interaction, role specialization, or response aggregation. Complex multi-agent systems, such as role-played debate or judge–advocate pipelines, may exhibit different or lower hallucination rates, but they also introduce additional variables and substantially higher deployment costs. Extending \textbf{LexAgentHallu} to such systems is an important direction for future work.

\section*{Acknowledgments}

This work is funded in part by the HKUST Start-up Fund (R9911), Theme-based Research Scheme grant (T45-205/21-N), the InnoHK initiative of the Innovation and Technology Commission of the Hong Kong Special Administrative Region Government, and the research funding under HKUST-DXM AI for Finance Joint Laboratory (DXM25EG01).


\bibliography{custom}

\appendix

\input{sections/app_ethics}

\input{sections/app_datasets}

\input{sections/app_annotation}

\input{sections/app_rubric_checklist}

\input{sections/app_judge}

\input{sections/app_exprimental_details}

\input{sections/app_detailed_analysis}

\input{sections/app_prompts_template}

\end{document}

%% file: sections/introduction.tex
The rapid progress of large language models (LLMs) has catalyzed a paradigm shift from static, single-turn generation to agentic reasoning, where models plan, invoke tools, and iteratively interact with their environment to solve complex tasks~\citep{yao2023react,anthropic2024building,xu2026theagentcompany,zheng2026skillprox,zhu2026selfevolvingdeepresearchjoint}. This shift is particularly attractive for high-stakes domains such as law, where reliable problem-solving demands statute retrieval, precedent grounding, and multi-hop reasoning that exceed the capacity of parametric memory alone. Reflecting this trend, recent efforts have actively pushed legal LLMs toward tool-augmented, multi-step legal agents, demonstrating clear gains on tasks such as legal question answering and judgement prediction~\citep{han2026trustworthy,han2026trustworthy1,zhang2025explicitsyllogisticlegalreasoning,yang2025glareagenticreasoninglegal,zhou2026lrasadvancedlegalreasoning,yang2026lawthinkerdeepresearchlegal}.



However, while the agentic paradigm alleviates certain single-turn hallucinations, it exposes new hallucination surfaces at the \emph{agentic} level. In legal agent systems, hallucinations may arise not only from legal reasoning such as fabricating case citations or misquoting statutory provisions, but also from the agentic process itself: the agent may invoke a tool with incorrect arguments, misinterpret a retrieved
passage, or lose track of the user's original request across turns. Understanding how often these failures occur and whether they stem from legal knowledge or agentic processes is essential for improving legal agent systems.~\citep{han-etal-2025-courtreasoner,lin2025llmbasedagentssufferhallucinations,zhou2026well}.

Several recent efforts have started to evaluate hallucinations in legal LLMs~\citep{Dahl_2024,hu-etal-2025-fine,han-etal-2025-courtreasoner} or in general-domain agents~\citep{liu2026agenthallubenchmarkingautomatedhallucination, zhu2025medinsightbenchevaluatingmedicalanalytics}. Despite these advances, no existing benchmark jointly examines to what extent legal agents hallucinate and at which level legal or agentic these hallucinations originate. We identify three critical limitations:

\noindent \textbf{(1) Lack of Agent-Level Hallucination Evaluation in the Legal Domain.} Existing legal hallucination benchmarks~\citep{hu-etal-2025-fine} are confined to single-turn QA over a narrow slice of legal tasks, treating the model as a closed-book oracle and leaving tool use, planning, and multi-turn interaction essentially unmeasured.

\noindent \textbf{(2) Fragmented and Coarse-Grained Hallucination Taxonomies.} General-domain agent hallucination studies adopt taxonomies (e.g., tool-call vs.\ output errors) that are blind to legally salient distinctions such as fabricated provisions and misapplied precedents. Conversely, legal hallucination taxonomies overlook agent-specific failures such as plan deviation and tool-grounding errors. Neither line of work offers a taxonomy that is simultaneously comprehensive and fine-grained enough to diagnose legal agent behavior.

\noindent \textbf{(3) Limited Error Attribution Across Knowledge and Process Layers.} Current evaluations score only the final answer with exact match or LLM-as-a-judge, without attributing errors to specific steps in the agent trajectory. When a legal agent produces a hallucinated conclusion, users cannot determine whether it stems from flawed legal knowledge or faulty agentic processes, limiting actionable diagnosis.

To bridge these gaps, we introduce \ourbench{}, the first benchmark for profiling hallucinations in legal LLM agents. \ourbench{} is constructed through an expert-in-the-loop pipeline, in which licensed legal professionals author and verify queries, gold trajectories, and rubric-based annotations across diverse legal tasks. We propose a dual-layer hallucination taxonomy that decomposes agentic legal errors into a \textbf{Substantive Layer} covering legal knowledge errors and an \textbf{Agentic Layer} covering process-level failures. To quantify behavior under this taxonomy, we design fine-grained metrics with step-level attribution, measuring to what extent and how a legal agent hallucinates along its trajectory.

In summary, our contributions are:
\begin{itemize}
    \item We release \ourbench{}, the first benchmark dedicated to hallucinations in legal LLM agents, featuring expert-curated queries, gold references, and taxonomy-driven rubric checklist across diverse legal tasks and categories.
    \item We propose a dual-layer taxonomy with fine-grained sub-classes that jointly capture what the agent gets wrong about the legal and agentic level during reasoning.
    \item We introduce taxonomy-aligned evaluation metrics including hallucination frequency, density, substantive/procedural cleanliness, and a Right-Answer-Wrong-Reason rate that go beyond outcome-only scoring to quantify how often and how broadly each rollout hallucinates, enabling diagnosis of failure patterns at both the layer and subclass level.
    \item  We benchmark a wide range of proprietary and open-source legal and general agents, uncovering systematic analysis invisible to outcome-level evaluation (details in \S\ref{sec:experiments}).
\end{itemize}

\input{table/legal_taxonomy}

%% file: table/legal_taxonomy.tex
\begin{table*}[t]
\centering
\setlength{\tabcolsep}{5pt}
\renewcommand{\arraystretch}{1}
\resizebox{0.7\textwidth}{!}{%
\begin{tabular}{@{}llll@{}}
\toprule
\multicolumn{1}{c}{\textbf{Top-layer}} & \multicolumn{1}{c}{\textbf{Mid-layer}} & \multicolumn{1}{c}{\textbf{Subclass}} & \multicolumn{1}{c}{\textbf{Description}} \\
\midrule
\multirow{19}{*}{\textcolor{bluedark}{\textbf{L1 Substantive}}}
 & \multirow{4}{*}{L1.1 Authority}
   & Source Fabrication            & Non-existent statute or case \\
 & & Citation--Content Misapplication & Real source, wrong application \\
 & & Hierarchy Error               & Confuses source rank \\
 & & Granularity Error             & Wrong paragraph or item \\
\cmidrule(l){2-4}
 & \multirow{6}{*}{L1.2 Doctrine}
   & Conceptual Confusion          & Conflates legal concepts \\
 & & Element Misstatement          & Distorts rule elements \\
 & & Exception Omission            & Ignores provisos \\
 & & Consequence Error             & Wrong legal effect \\
 & & Doctrinal Position Confusion  & Mixes doctrinal stances \\
 & & Discretionary-Judgment Error  & Misapplies discretion \\
\cmidrule(l){2-4}
 & \multirow{5}{*}{L1.3 Procedural Law}
   & Jurisdiction Error            & Wrong court or venue \\
 & & Period Error                  & Wrong deadline or limit \\
 & & Procedural-Step Error         & Skips or misorders steps \\
 & & Procedural-Outcome Error      & Wrong disposition \\
 & & Appeal Error                  & Wrong remedy path \\
\cmidrule(l){2-4}
 & \multirow{4}{*}{L1.4 Application \& Subsumption}
   & Fact Fabrication              & Invents case facts \\
 & & Fact Omission                 & Drops material facts \\
 & & Element--Fact Mismatch        & Maps facts to wrong elements \\
 & & Party Confusion               & Mixes up party roles \\
\midrule
\multirow{9}{*}{\textcolor{reddark}{\textbf{L2 Agent-Proc.}}}
 & \multirow{5}{*}{L2.1 Planning \& Reasoning}
   & Premature Closure             & Concludes too early \\
 & & Syllogism Error               & Invalid inference \\
 & & Self-Contradiction            & Inconsistent claims \\
 & & Step Skip / Conflation        & Merges distinct steps \\
 & & Out-of-Context Quoting        & Cite used out of context \\
\cmidrule(l){2-4}
 & \multirow{1}{*}{L2.2 Memory}
& Memory Hallucination          & Multi-turn forgetting and misremembering \\
\cmidrule(l){2-4}
 & \multirow{2}{*}{L2.3 Tool-Call \& Observation}
   & Tool-Call Error               & Wrong tool or arguments \\
 & & Observation Misuse            & Misreads tool output \\
\bottomrule
\end{tabular}%
}
\caption{Hierarchical taxonomy of hallucinations (2 layers, 7 mid-level layers, 27 fine-grained classes).}
\label{tab:taxonomy_tree}
\end{table*}

%% file: sections/related_work.tex
\subsection{LLM Agents and Legal Agents}

A first wave of legal LLMs, including \textit{LawGPT}~\citep{lawgpt}, \textit{ChatLaw}~\citep{cui2024chatlaw}, \textit{Lawyer-LLaMA}~\citep{huang2023lawyer}, \textit{DISC-LawLLM}~\citep{yue2023disclawllm}, and \textit{LexiLaw}~\citep{LexiLaw}, adapted general models to legal corpora via continued pre-training and instruction tuning, but remain closed-book and single-turn. Agentic frameworks such as \textit{ReAct}~\citep{yao2023react} and \textit{Plan-and-Execute}~\citep{topsakal2023creating} extend LLMs into autonomous problem solvers that decompose tasks, call external tools, and refine intermediate results. Building on this paradigm, \textit{LawThinker}~\citep{yang2026lawthinkerdeepresearchlegal} and \textit{LRAS}~\citep{zhou2026lrasadvancedlegalreasoning} equip LLMs with search tools. While these systems expand the practical utility of LLMs in law, their reliability under multi-step, tool-augmented execution remains largely uncharacterized.

\subsection{Legal Benchmarks}
A wide range of benchmarks evaluate LLMs on legal tasks. \textit{LawBench}~\citep{fei2024lawbench} and \textit{LexEval}~\citep{lexeval} cover broad task suites in Chinese law; \textit{LegalBench}~\citep{guha2023legalbench} targets common-law reasoning; \textit{PLawBench}~\citep{shi2026plawbench} and \textit{J1-Eval}~\citep{J1} extend to procedural law and judicial reasoning; and \textit{LEXam}~\citep{fan2025lexam} focuses on bar-exam-style multi-step questions. Despite their breadth, these benchmarks target task accuracy in single-turn, closed-book settings, leaving hallucination in multi-step, tool-augmented agentic workflows unexplored.

\subsection{Hallucination Benchmarks}
General hallucination benchmarks such as \textit{HaluEval}~\citep{li-etal-2023-halueval}, \textit{TruthfulQA}~\citep{lin2022truthfulqameasuringmodelsmimic}, \textit{FActScore}~\citep{min-etal-2023-factscore}, and \textit{FELM}~\citep{zhao2023felm} pioneered evaluation of factual errors in LLM outputs. In the legal domain, \textit{LegalHalBench}~\citep{hu-etal-2025-fine} and \textit{CitaLaw}~\citep{zhang2025citalaw} extend hallucination evaluation to statutes and citations. However, all of these are confined to single-turn outputs, treating hallucination as a property of an isolated answer rather than a multi-step trajectory. More recently, \textit{AgentHalluBench}~\citep{liu2026agenthallubenchmarkingautomatedhallucination}, \textit{HaluAgent}~\citep{cheng2024small}, and \textit{ToolBH}~\citep{zhang-etal-2024-toolbehonest} evaluate hallucinations in tool-using agents, but target generic domains with task-agnostic taxonomies and lack legal grounding. Consequently, no existing benchmark evaluates hallucinations in multi-step, tool-augmented legal workflows spanning both substantive and procedural errors.

\begin{figure*}[t]
    \centering
    \includegraphics[width=0.9\linewidth]{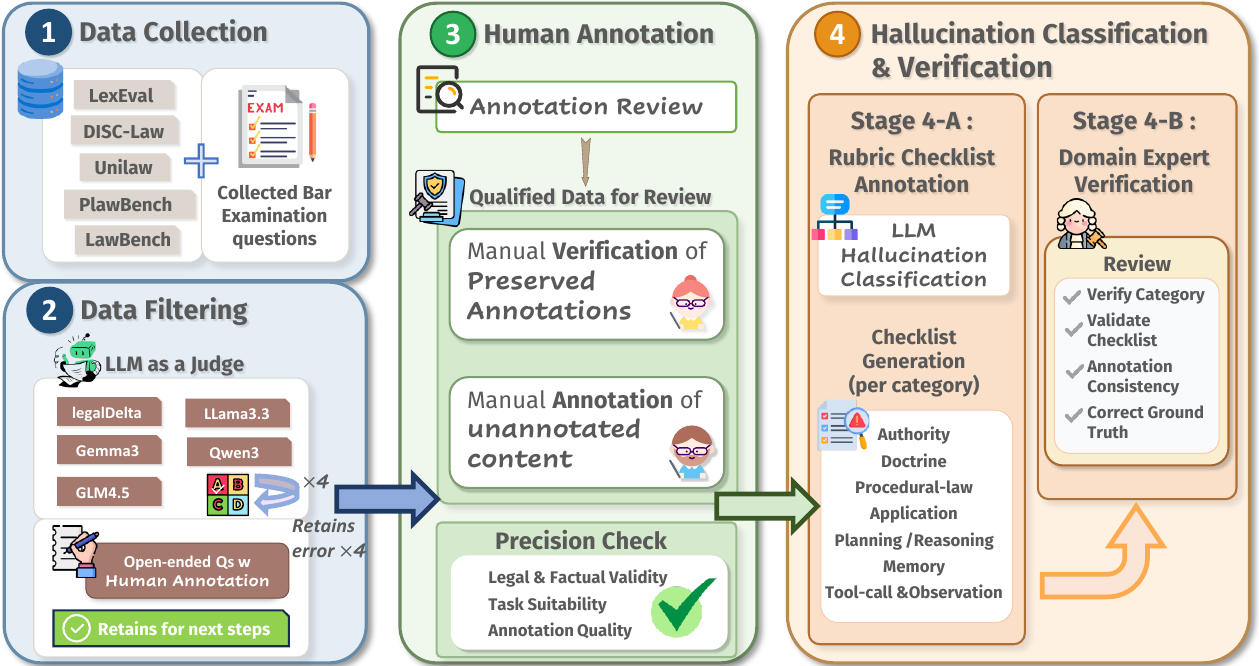}
    \caption{The four-stage pipeline for constructing \textsc{\ourbench}.}
    \label{fig:pipeline}
\end{figure*}

%% file: sections/benchmark.tex

In this section, we introduce \textsc{\ourbench}, an expert-curated benchmark for diagnosing hallucinations in legal LLM-based agents. Existing legal benchmarks predominantly score end-task accuracy, conflating what an agent answers with how it arrives there. For autonomous agents, however, a correct answer can be reached through an unsound procedure, and a well-formed trajectory can still yield a wrong conclusion. \textsc{\ourbench} is built around this distinction via three design choices: (i)~a two-layer hallucination taxonomy jointly characterizing substantive legal errors and agent-procedural errors (\S\ref{sec:taxonomy}); (ii)~a four-stage human-in-the-loop curation pipeline concentrating evaluation on hard cases with verified ground truth (\S\ref{sec:curation}); and (iii)~a taxonomy-anchored rubric checklist that converts ground-truth answers into fine-grained evaluation signals for an LLM-as-a-judge system (\S\ref{sec:judge}). We describe each component below, followed by dataset statistics (\S\ref{sec:stats}).
\vspace{-2mm}
\subsection{Dual-Layer Hallucination Taxonomy}
\label{sec:taxonomy}
\textsc{\ourbench} evaluates hallucinations along two orthogonal dimensions: \textbf{Layer~1} captures errors in what the agent says about the law, and \textbf{Layer~2} captures errors in how it arrives at that statement. Prior legal hallucination studies address only the former at coarse granularity~\citep{Dahl_2024,hu-etal-2025-fine,han-etal-2025-courtreasoner}, while general agentic-evaluation work~\citep{lin2025llmbasedagentssufferhallucinations,liu2026agenthallubenchmarkingautomatedhallucination} addresses only the latter in a domain-agnostic manner; neither alone suffices for legal agents. Treating the two layers as orthogonal allows a single trajectory step to incur hallucinations from one layer, the other, both, or neither, letting \textsc{\ourbench} attribute each error to both a legal-content failure mode and the agentic mechanism responsible. Developed with licensed legal practitioners over iterative consolidation rounds, the taxonomy comprises \textbf{7 mid-level categories} and \textbf{27 fine-grained subclasses} (Table~\ref{tab:taxonomy_tree}). Full definitions and worked examples are in Appendix~\ref{app:app_tax}.

\subsection{Curation of Benchmark}
\label{sec:curation}


To make \textsc{\ourbench} realistic in difficulty, comprehensive in legal coverage, and faithfully aligned with the taxonomy, we curate it through a four-stage pipeline as shown in the Figure~\ref{fig:pipeline}
Each stage is paired with a dedicated human-in-the-loop step described below.

\noindent \textbf{Stage 1: Multi-source data collection.} Legal-agent evaluation requires both broad coverage of standardized academic tasks and exposure to up-to-date professional practice, neither of which is sufficient alone. We therefore aggregate raw questions from two complementary streams. The open-source stream draws from four widely used Chinese legal benchmarks, \textit{LexEval}~\citep{lexeval}, \textit{LawBench}~\citep{fei2024lawbench}, \textit{UniLaw}~\citep{cai-etal-2025-unilaw}, \textit{DISC-Law-Eval}~\citep{yue2023disclawllmfinetuninglargelanguage}, and \textit{PLawBench}~\citep{shi2026plawbench}. The real-world-practice stream consists of items manually collected from China's National Judicial Examination (\textit{Fakao}) released in the most recent three years. Detailed descriptions of each constituent dataset and licensing, are provided in Appendix~\ref{app:datasets}.

\noindent \textbf{Stage 2: Data Filtering.} Given the heterogeneity of the collected sources, a single filtering rule would either discard too much open-ended supervision or retain too many easy closed-ended items. We therefore apply task-type-specific strategies that distinguish closed-ended questions from open-ended questions.


For closed-ended questions, we use a multi-model rollout filter to remove items that current LLMs can already solve reliably. Each candidate is rolled out four times by each of six diverse models: \textit{LegalDelta-4B}, \textit{LegalDelta-14B}~\citep{gemmateam2025gemma3technicalreport}, \textit{Qwen3-30B-A3B}~\citep{qwen3}, \textit{Gemma3-4B}~\citep{gemmateam2025gemma3technicalreport}, \textit{Llama3.3-70B}~\citep{llama3herdmodels}, and \textit{GLM-4.5-Flash}~\citep{GLM4.5}. We retain only questions on which all rollouts from every model fail, yielding a hard-case subset where hallucination is empirically likely.
For open-ended questions, filtering depends on available supervision. \textit{PLawBench} items are retained in full as they already carry expert-authored rubric annotations. 

For open-ended subsets of \textit{LexEval} and \textit{LawBench}, we apply an LLM-as-a-judge filter that scores candidates along multiple legal-quality dimensions, retaining only items receiving the highest score across all dimensions. Judge prompts and scoring rubrics are reported in Appendix~\ref{app:app_prompts}.

\noindent \textbf{Stage 3: Expert Annotation and Verification.} After Stage~2, the filtered pool contains both items with pre-existing expert annotations and items lacking ground-truth supervision. We therefore apply a two-track annotation protocol carried out by annotators with formal legal training. \emph{(a)~Verification track:} For items already carrying source-benchmark annotations, annotators verify each ground-truth answer against authoritative statutory and doctrinal sources, correcting inaccuracies and discarding items whose supervision cannot be reliably reconstructed. \emph{(b)~De novo track:} For items without annotations (primarily from the Judicial Examination set), annotators produce gold answers, supporting statutory citations, and expected reasoning trajectories. Each item is independently labeled by two annotators.

\noindent \textbf{Stage 4: Taxonomy-driven rubric checklist.} A verified ground-truth answer alone is a coarse reference signal: it tells an LLM-as-a-judge whether the agent is right, but not which taxonomy subclass it violates when wrong, nor at which step the violation occurs. To bridge this gap, we introduce a \emph{taxonomy-driven rubric checklist} as a distinguishing feature of \textsc{\ourbench}. Concretely, taking the question and its verified ground-truth answer as input, we prompt \textit{Claude-4.6-Sonnet} to generate a fine-grained checklist in which each item is explicitly anchored to a hallucination subclass defined in Layer 1 of the taxonomy (\S\ref{sec:taxonomy}). Each generated checklist is then \emph{human-verified}: legal experts inspect every item for factual correctness, taxonomy alignment, and non-redundancy, revising or removing items where necessary. A representative example is shown in Appendix~\ref{app:app_tax} and the full generation prompt in Appendix~\ref{app:app_prompts}. The resulting rubric checklists form the backbone of our judge system (\S\ref{sec:judge}), enabling interpretable, taxonomy-aligned hallucination diagnosis.

\subsection{Dataset Statistics}
\label{sec:stats}

\begin{figure}[t]
    \centering
    \includegraphics[width=0.9\linewidth]{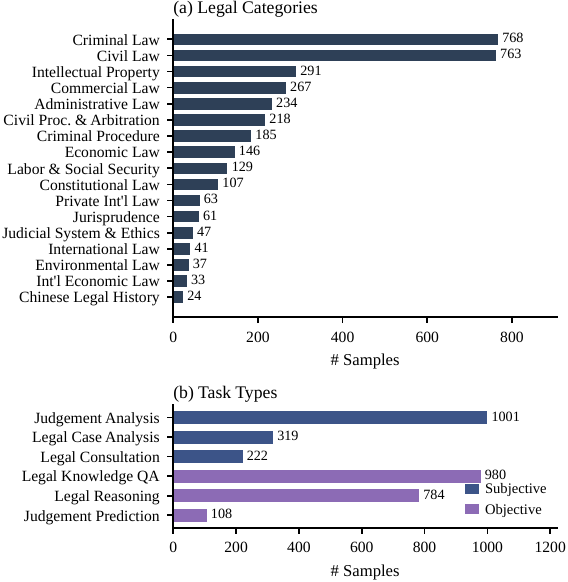}
    \caption{Distribution of \textsc{\ourbench} instances across (a) 17 legal categories and (b) 6 task types.}
    \label{fig:stats}
\end{figure}

\begin{figure}[t]
    \centering
    \includegraphics[width=0.9\linewidth]{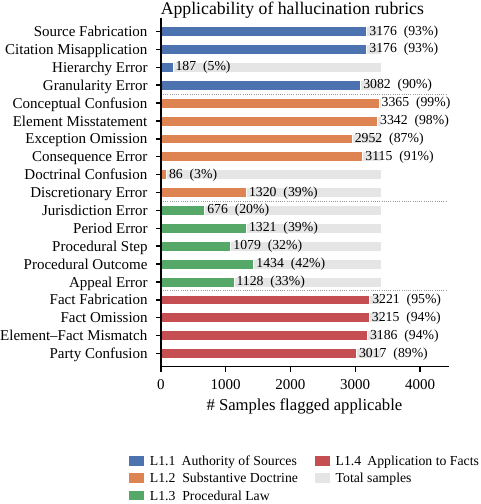}
    \caption{Applicability of rubric checklist items across Layer-1 hallucination subclasses, colored by top-level category.}
    \label{fig:rubric}
\end{figure}

\textsc{\ourbench} comprises \textbf{3,414} expert-curated instances spanning \textbf{17} legal categories and \textbf{6} task types.  As shown in Figure~\ref{fig:stats}(a), the datasets anchored in criminal and civil law, with broad coverage spanning intellectual property, commercial, administrative, constitutional and international law, as well as legal history. Figure~\ref{fig:stats}(b) breaks down instances by task type: 1,542 (45.2\%) are subjective and 1,872 (54.8\%) are objective.

Each instance is paired with a taxonomy-anchored rubric checklist (\S\ref{sec:curation}, Stage4). Figure\ref{fig:rubric} summarizes the applicability of rubric items across all 19 Layer-1 subclasses. Application-to-Facts subclasses apply to over 89\% of samples, reflecting the universal need for fact-based reasoning. Most Authority and Substantive-Doctrine subcalsses display comparably high applicability, with a few narrowly scoped exceptions. Procedural-Law subclasses are more context-dependent, ranging from broadly relevant (procedural-outcome, period errors) to narrowly invoked (jurisdiction errors).

\subsection{Our Judge System}
\label{sec:judge}

\subsubsection{How to Judge.}

Since every rubric item is anchored to a specific taxonomy subclass
(\S\ref{sec:curation}, Stage~4), evaluating an agent's output reduces 
to a per-item decision: whether the corresponding hallucination subclass 
is present in the trajectory. We implement this as a \emph{rubric-anchored LLM-as-a-judge} procedure with seven dedicated prompts---one per top-level taxonomy category (L1.1--L1.4 and L2.1--L2.3)---so that each call focuses the judge on a narrow, coherent error family rather than the full 27-subclass space. Given an agent trajectory $r$ (comprising the final answer together with intermediate reasoning, memory states, and tool-call observations) and the instance's rubric checklist, the judge returns a binary verdict per rubric item along with the trajectory span responsible for any flagged violation. Crucially, the judge operates on the \emph{full trajectory rather than the final answer alone}; this is essential for detecting Layer-2 errors. We then aggregate per-item verdicts into the rollout-level metrics defined below. Prompt templates and robustness analysis are provided in Appendix~\ref{app:app_prompts} and Appendix~\ref{app:judge_robustness}.

\subsubsection{Metric design.}



\noindent \textbf{Hallucination Frequency.} Hallucination Frequency (HF) measures whether at least one hallucination occurs within a target rollout, aggregated across rollouts. 
The overall and layer-level hallucination frequency are defined as follows:

\begin{equation}
\begin{aligned}
HF &= \frac{1}{N}\sum_{r=1}^N \mathbb{I}\!\left(\sum_s h_{r,s}\geq 1\right), \\
HF_{L\ell} &= \frac{1}{N}\sum_{r=1}^N \mathbb{I}\!\left(\sum_{s\in L\ell} h_{r,s}\geq 1\right),
\end{aligned}
\label{eq:hf}
\end{equation}
where $\ell\in\{1,2\}$ denotes the taxonomy level, $\mathbb{I}(\cdot)$ is the indicator function, $N$ is the number of rollouts, and $h_{r,s}\in\{0,1\}$ indicates whether rollout $r$ triggers hallucination subclass $s$.



\noindent \textbf{Hallucination Density.} Hallucination Density normalizes the number of triggered subclasses by the number of applicable ones, yielding a per-rollout severity score in $[0,1]$ averaged over the corpus. This design directly addresses a key limitation of HF, which is insensitive to multiplicity: under HF, rollouts violating one or ten subclasses contribute identically. Thus, the layer-level Hallucination Density is formulated by

\begin{equation}
\begin{aligned}
HD_{L\ell} = \frac{1}{N}\sum_{{r=1}}^N \frac{\sum_{s\in {{L_\ell}}}}{N_{{L \ell(r)}}}h_{r,s},
\end{aligned}
\label{HD}
\end{equation}

\noindent where $N_{{L \ell(r)}}$ is the number of categories in the $\ell$ layer. Additional metrics---including Answer Correctness, Substantive/Procedural Cleanliness, the Right-Answer-Wrong-Reason (RAWR) rate, and the co-occurrence Lift matrix used in our analysis---are formally defined in Appendix~\ref{app:judge}.

%% file: sections/experiments.tex
\subsection{Experimental Setup}


To comprehensively assess legal agentic behavior, we evaluate two categories of systems on \textsc{\ourbench}. The first comprises the \textsc{LRAS} family (4B, 8B, 14B), standalone legal agent models running their native workflows. The second isolates the effect of orchestration: we pair three agentic frameworks (\textsc{LawThinker}, Plan-and-Execute, and ReAct) with a shared legal tool suite and a range of backbones, including open-source models, including Qwen3.5-9B, Qwen3.5-27B~\citep{qwen3}, and Qwen3.6-27B~\citep{qwen3.6-27b}) and closed-source models, such as Gemini-3.1-Pro, and GPT-5.4~\citep{singh2026openaigpt5card}. This design enables direct comparison between specialized legal agents and general backbones under controlled tool access. Detailed setups are in Appendix~\ref{app:app_exp_set}.

\input{table/main_results}
\subsection{Main Results}

Table~\ref{tab:main_leaderboard} summarizes the overall hallucination performance. We find that:

\noindent \textbf{1) Pervasive Hallucination Across All Systems.} Whether built as a general-purpose agentic workflow (Plan-and-Execute, ReAct), a legal-specialized workflow (LawThinker), or a dedicated legal agent model (LRAS), each configuration exhibits substantial hallucination rates. Even the best-performing entry, Gemini-3.1-Pro under ReAct, still triggers at least one hallucination in $89.0\%$ of rollouts (HF$=$0.890), with $\text{HF}_{L1}$ never dropping below $0.875$ for any model. This confirms that even frontier legal agents remain far from faithful legal reasoning.

\noindent \textbf{2) Framework-Dependent Hallucination Profiles.} Model rankings shift markedly across frameworks, indicating that no single orchestration strategy universally dominates. ReAct delivers the strongest suppression on overall and substantive hallucinations, with Gemini-3.1-Pro achieving the lowest HF ($0.890$) and Qwen3.6-27B the lowest $\text{HF}_{L1}$ ($0.875$). LawThinker excels at suppressing procedural ($L2$) hallucinations, where GPT-5.4 attains the best $\text{HF}_{L2}$ ($0.462$) and $\text{HD}_{L2}$ ($0.172$), but at the cost of elevated $L1$ rates. Plan-and-Execute yields uniformly mediocre results without winning any column. Finer-grained category and task profiles are provided in Section~\ref{sec:dec_analysis}.

\noindent \textbf{3) Scaling Behavior and Evaluation Stability.} Performance correlates positively with model scale: within the Qwen family under LawThinker, $3.5\text{-}9\text{B}\!\to\!3.5\text{-}27\text{B}\!\to\!3.6\text{-}27\text{B}$ monotonically reduces $\text{HF}_{L2}$ from $0.931$ to $0.754$. Frontier closed-source models still outperform the largest open-source variants on most metrics, exposing a persistent gap in open-source legal faithfulness. LRAS narrows this gap with substantially smaller backbones: LRAS-Qwen3-14B reaches $\text{HF}_{L2}$$=$$0.580$ and $\text{HD}_{L2}$$=$$0.231$, surpassing all ReAct and Plan-and-Execute configurations of frontier models on $\text{HD}_{L2}$, while even LRAS-Qwen3-4B outperforms the 27B Qwen baselines on $\text{HF}_{L2}$. Notably, all five metrics decrease monotonically from 4B to 8B to 14B, confirming that hallucination suppression scales favorably with model size and validating the discriminative power of our evaluation framework.

\subsection{Detailed Analysis}\label{sec:dec_analysis}

\begin{figure}[t]
    \centering
    \includegraphics[width=0.9\linewidth]{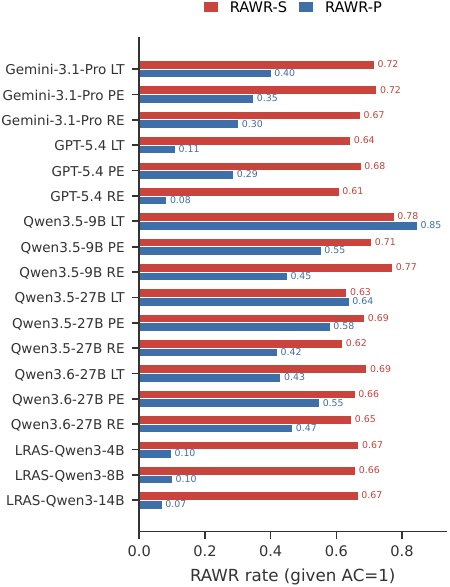}
    \caption{\textbf{RAWR rates on the objective-question subset.} Substantive (RAWR-S, red) and procedural (RAWR-P, blue) RAWR rates, conditional on a correct answer (AC=1). LT, PE, RE denote \textsc{LawThinker}, \textsc{Plan-and-Execute}, and \textsc{ReAct} frameworks. Metric definitions are in Appendix~\ref{app:judge}.}
    \label{fig:rawr}
\end{figure}


\noindent \textbf{Right-Answer-Wrong-Reason: a correct answer rarely implies a clean reasoning process.} As shown in Figure~\ref{fig:rawr}, even when a legal agent produces a correct final answer, its reasoning trajectory still contains at least one substantive-legal hallucination (RAWR-S$=68\%$) and at least one agent-procedural hallucination (RAWR-P$=37\%$) on average. RAWR-S is persistently high and varies only within a narrow band ($0.61$--$0.78$), from GPT-5.4 RE to Qwen3.5-9B LT, and even the constrained LRAS remain at $0.66$--$0.67$, confirming that substantive-legal errors are largely insensitive to model scale, backbone, or agentic framework. In contrast, RAWR-P varies much more ($0.07$--$0.85$) and shows two patterns: (i) a framework effect, where LRAS keeps agent-procedural hallucinations at $\leq0.10$ for all backbones, while multi-tool frameworks leave open-source models in the $0.42$--$0.85$ range; (ii) a backbone effect, where GPT-5.4 reaches LRAS-level cleanliness even with multi-tool frameworks (LT: $0.11$; RE: $0.08$), while Qwen3.5-9B LT peaks at $0.85$. ReAct achieves the lowest RAWR-P among multi-tool frameworks across all backbones, suggesting that action-observation grounding helps limit reasoning drift. These results demonstrate that objective-question accuracy is an unreliable proxy for legal faithfulness: even the best-performing configurations exhibit over $60\%$ substantive contamination in their reasoning trajectories.

\begin{figure}[t]
    \centering
    \includegraphics[width=0.9\linewidth]{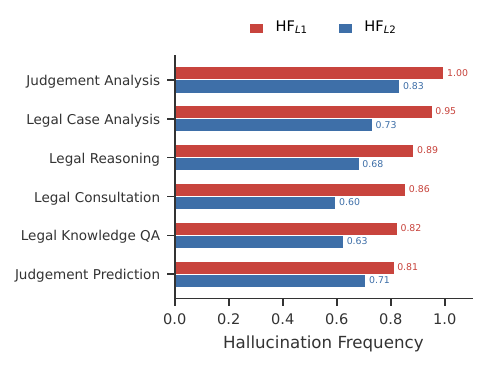}
    \caption{Hallucination frequency by task type, sorted by HF$_{L1}$ descending.}
    \label{fig:task_difficulty}
\end{figure}

\noindent \textbf{Task-Type and Legal-Category Profiles.} As shown in Figure~\ref{fig:task_difficulty}, open-ended generation tasks such as Adjudication Analysis and Case Analysis saturate $\text{HF}_{L1}$ near $1.00$, while constrained tasks such as Legal Knowledge QA ($0.82$) and Judgement Prediction ($0.81$) sit lower, likely because shorter outputs limit the surface area for hallucination. $\text{HF}_{L2}$ follows a different ordering: Adjudication Analysis still leads ($0.83$), yet Judgement Prediction ($0.71$) surpasses Legal Reasoning ($0.68$) and Legal Consultation ($0.60$), suggesting that multi-step evidence aggregation stresses planning and memory even when the final output is short. 
At the legal-category level (Figure~\ref{fig:app_legal_category}), procedural and core doctrinal branches such as Criminal Procedure ($0.95) and Civil Law ($0.93) form a high-hallucination cluster, while theory-oriented categories such as Constitutional Law ($0.79$) and Jurisprudence ($0.65$) rank lowest on both layers. This likely reflects a structural difference in legal reasoning. Procedural and doctrinal fields require precise rule selection, source hierarchy control, deadline computation, and fact--rule application, whereas theory-oriented categories more often involve abstract principles and conceptual explanation, leaving fewer points at which a model can misstate a concrete provision, procedural step, or factual predicate~\citep{linna2026challenges,fan2025lexam}.

\begin{figure}[t]
    \centering
    \includegraphics[width=0.9\linewidth]{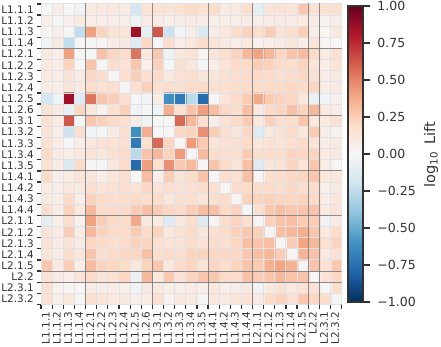}
    \caption{\textbf{27$\times$27 subclass co-occurrence matrix.} Cell colour is $\log_{10}\text{Lift}$ as defined in \S\ref{sec:judge}; red indicates positive co-occurrence, blue negative, white near independence. Black lines mark group boundaries.}
    \label{fig:cooccurrence}
\end{figure}

\noindent \textbf{Hallucination subclasses cluster, not scatter.} The lift matrix in Figure~\ref{fig:cooccurrence} reveals structure invisible to per-rollout aggregates. First, within-layer coupling dominates: dark-red blocks concentrate along the L1.2 and L1.3 diagonals, showing that once one doctrine or procedural error appears it reliably triggers further errors of the same family. Second, specific mechanistic pairs rise sharply above chance co-occurrence. Within L1.3, Temporal/Limitation errors co-occur with Remedy-Path errors (Lift$\,=2.82$) and Procedural-Step with Procedural-Consequence errors (Lift$\,=2.59$), suggesting a shared procedural clock whose miscalibration simultaneously corrupts deadlines and the remedies gated by them. The strongest cross-group signal is L1.1.3$\,\leftrightarrow\,$L1.2.5 (Lift$\,=7.07$): legal-hierarchy errors and doctrinal-stance conflation systematically co-occur, pointing to a common upstream failure in mapping source hierarchy to doctrinal position. These findings indicate that legal hallucinations are not independent slips but structured failure modes rooted in shared mechanisms, a property that aggregate accuracy or per-error rates cannot expose. Practically, this co-occurrence structure suggests that detecting high-frequency hallucination types can serve as early signals to proactively flag and mitigate their correlated low-frequency counterparts before they propagate.


%% file: table/main_results.tex
\begin{table}[t]
\centering
\small
\setlength{\tabcolsep}{3pt}
\renewcommand{\arraystretch}{1.2}
\begin{tabular}{lccccc}
\toprule
\textbf{Model} & \textbf{HF}$\downarrow$ & \textbf{HF}$_{L1}$ & \textbf{HF}$_{L2}$ & \textbf{HD}$_{L1}$ & \textbf{HD}$_{L2}$ \\
\midrule
\multicolumn{6}{c}{\cellcolor{gray!15}\textit{LawThinker}} \\
Gemini-3.1-Pro  & 0.911 & 0.881 & 0.627 & \textbf{0.258} & 0.233 \\
GPT-5.4         & 0.909 & 0.900 & \textbf{0.462} & 0.279 & \textbf{0.172} \\
Qwen3.5-9B     & 0.987 & 0.944 & 0.931 & 0.355 & 0.470 \\
Qwen3.5-27B    & 0.956 & 0.891 & 0.860 & 0.304 & 0.375 \\
Qwen3.6-27B    & 0.940 & 0.906 & 0.754 & 0.301 & 0.327 \\
\midrule
\multicolumn{6}{c}{\cellcolor{gray!15}\textit{Plan-and-Execute}} \\
Gemini-3.1-Pro  & 0.899 & 0.882 & 0.643 & 0.279 & 0.254 \\
GPT-5.4         & 0.920 & 0.905 & 0.666 & 0.293 & 0.256 \\
Qwen3.5-9B     & 0.953 & 0.921 & 0.855 & 0.330 & 0.356 \\
Qwen3.5-27B    & 0.945 & 0.906 & 0.844 & 0.314 & 0.335 \\
Qwen3.6-27B    & 0.933 & 0.892 & 0.829 & 0.303 & 0.315 \\
\midrule
\multicolumn{6}{c}{\cellcolor{gray!15}\textit{ReAct}} \\
Gemini-3.1-Pro  & \textbf{0.890} & 0.879 & 0.617 & 0.279 & 0.247 \\
GPT-5.4         & 0.909 & 0.903 & 0.601 & 0.302 & 0.242 \\
Qwen3.5-9B     & 0.957 & 0.944 & 0.823 & 0.345 & 0.345 \\
Qwen3.5-27B    & 0.913 & 0.877 & 0.739 & 0.291 & 0.251 \\
Qwen3.6-27B    & 0.916 & \textbf{0.875} & 0.742 & 0.286 & 0.251 \\
\midrule
\multicolumn{6}{c}{\cellcolor{gray!15}\textit{LRAS}} \\
LRAS-Qwen3-4B  & 0.924 & 0.922 & 0.614 & 0.324 & 0.249 \\
LRAS-Qwen3-8B  & 0.914 & 0.912 & 0.600 & 0.319 & 0.238 \\
LRAS-Qwen3-14B & 0.911 & 0.909 & 0.580 & 0.311 & 0.231 \\
\bottomrule
\end{tabular}
\caption{Overall hallucination performance. All metrics are lower-is-better; best per column in \textbf{bold}.}
\label{tab:main_leaderboard}
\end{table}

%% file: sections/conclusion.tex
We presented \textsc{LexAgentHallu}, the first benchmark for evaluating to what extent and how legal LLM agents hallucinate along multi-step trajectories. Built through a four-stage expert-in-the-loop pipeline, \textsc{LexAgentHallu} provides 3{,}414 instances across 17 legal categories and 6 task types, each annotated under a dual-layer taxonomy of 7 mid-level categories and 27 fine-grained subclasses covering both substantive and agentic failures. Our fine-grained metrics enable step-level attribution that localizes where each failure occurs along an agent's execution path. Evaluation of 18 proprietary and open-source agents uncovers a Right-Answer-Wrong-Reason effect and reveals that hallucination subclasses cluster rather than scatter, forming distinct agentic framework, legal task, and category profiles. These findings, invisible to outcome-level evaluation, validate the diagnostic power of \textsc{LexAgentHallu} and argue for a shift toward trajectory-aware evaluation in legal agent design. We anticipate that \textsc{LexAgentHallu} will serve as a foundation for developing more trustworthy legal agents.

%% file: sections/app_ethics.tex
\section{Ethics Statement}
\label{sec:app_ethic}

All data used in this work originates from publicly available or properly licensed sources, and we have verified compliance with the respective licensing terms. The dataset does not contain any personally identifiable or sensitive information. Human participation was limited to annotation and quality verification; all annotators are domain experts holding qualifications in law or and were compensated at a fair hourly rate commensurate with their expertise.

%% file: sections/app_datasets.tex
\section{Datasets}
\label{app:datasets}

This appendix provides detailed descriptions of the five legal benchmarks used in curation of \ourbench. Together, they span a broad spectrum of legal cognitive demands—from factual recall and concept recognition to multi-step reasoning and practical document drafting—enabling a comprehensive assessment of hallucination patterns across diverse task types.

\subsection{LexEval}
\label{app:lexeval}

LexEval~\citep{lexeval} is a large-scale Chinese legal evaluation suite that organizes 23 tasks (approximately 14,150 questions) under the Legal Cognitive Ability Taxonomy (LexCog). Tasks are sourced from established legal corpora, real bar-examination items, and expert-curated annotations. Beyond standard legal knowledge assessment, LexEval uniquely incorporates ethical reasoning scenarios, testing whether models can navigate value conflicts that arise in legal practice. We select a representative subset covering knowledge recall, statute interpretation, and case-based reasoning tasks.

\subsection{LawBench}
\label{app:lawbench}

LawBench~\citep{fei2024lawbench} structures its evaluation around three cognitive levels that mirror progressive stages of legal expertise: \emph{Memorization} (retrieval of statutory provisions and doctrinal facts), \emph{Understanding} (entity recognition, relation extraction, and semantic comprehension), and \emph{Application} (multi-step reasoning over realistic legal scenarios). Its 20 tasks adopt five output formats—single-label classification, multi-label classification, regression, extraction, and generation—providing diversity in both cognitive demand and answer granularity.

\subsection{UniLaw-Eval}
\label{app:unilaw}

UniLaw-Eval~\citep{cai-etal-2025-unilaw} targets logical inference within legal contexts through 800 rigorously constructed items (426 single-choice and 374 multi-choice questions). Each item is designed to require multi-step deductive or analogical reasoning rather than surface-level pattern matching, making it particularly suitable for probing doctrinal and subsumption hallucinations (L1.2 and L1.4 in our taxonomy).

\subsection{DISC-LawEval}
\label{app:disclaw}

DISC-LawEval~\citep{yue2023disclawllm} adopts a dual-track evaluation paradigm. Its \emph{objective} track draws multiple-choice questions from standardized professional examinations (e.g., the National Unified Legal Profession Qualification Examination) and stratifies them into three difficulty tiers to differentiate knowledge retrieval from deep deduction. Its \emph{subjective} track provides 300 expert-constructed open-ended scenarios—including legal consultation and judgment prediction—assessed along accuracy, completeness, and clarity dimensions, offering a natural testbed for detecting reasoning and application hallucinations.

\subsection{PLawBench}
\label{app:plawbench}

PLawBench~\citep{shi2026plawbench} is a practice-oriented benchmark designed to bridge the gap between academic evaluation and real-world legal workflows. It models three core practitioner activities: public legal consultation, practical case analysis, and legal document generation. The benchmark comprises 850 questions spanning 13 practice scenarios, each accompanied by expert-designed multi-dimensional rubrics (approximately 12,500 rubric items in total). These fine-grained rubrics assess issue identification, fact extraction, structured reasoning, and document coherence, making PLawBench particularly well-suited for evaluating whether models produce legally sound and internally consistent outputs under realistic task complexity.

\vspace{0.5em}
\noindent\textbf{Summary statistics.} Table~\ref{tab:dataset_stats} provides an overview of the scale, task format, and primary cognitive focus of each benchmark.

\begin{table}[t]
\centering
\small
\setlength{\tabcolsep}{6pt}
\renewcommand{\arraystretch}{1.1}
\begin{tabular}{@{}lccc@{}}
\toprule
\textbf{Benchmark} & \textbf{\#Items} & \textbf{\#Tasks} & \textbf{Format} \\
\midrule
LexEval      & $\sim$14,150 & 23 & Mixed \\
LawBench     & --           & 20 & Mixed \\
UniLaw-Eval  & 800          & -- & MCQ \\
DISC-LawEval & --           & -- & MCQ+Open \\
PLawBench    & 850          & 13 & Open \\
\bottomrule
\end{tabular}
\caption{Overview of evaluation benchmarks.}
\label{tab:dataset_stats}
\end{table}

\begin{figure*}[h]
    \centering
    \includegraphics[width=0.92\linewidth]{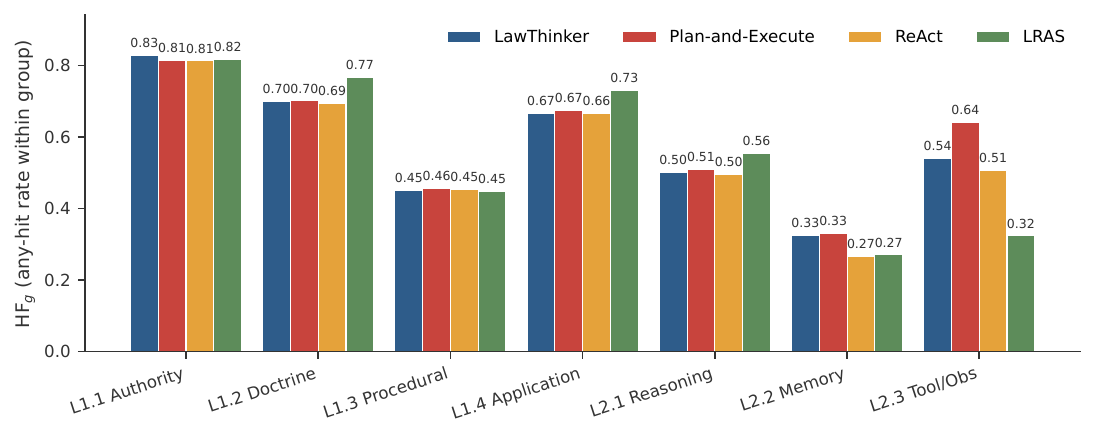}
    \caption{Per-group hallucination profile across the four agent
    frameworks.  Bars are HF$_g$ (any-hit rate within group).}
    \label{fig:app_framework_spec}
\end{figure*}

\begin{figure}[h]
    \centering
    \includegraphics[width=0.92\linewidth]{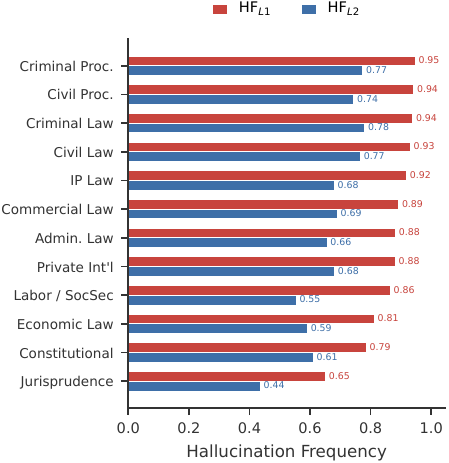}
    \caption{Hallucination frequency for the top-12 legal categories in \textsc{\ourbench}, sorted by HF$_{L1}$ descending. Procedural and core doctrinal branches (Criminal Proc., Civil Proc., Criminal Law, Civil Law) cluster above $0.93$, while theory-oriented categories (Constitutional, Jurisprudence) sit notably lower.}
    \label{fig:app_legal_category}
\end{figure}

%% file: sections/app_annotation.tex
\section{Annotation Details}
\label{app:annotation}

Two experts carried out all annotation work over a period of 21 days, each contributing an average of 4--6 hours per day. For instance, without
pre-existing ground truth, the annotators produced gold-standard labels from scratch. In instances where ground truth was already available, the annotators verified it, confirming correctness or correcting minor discrepancies. The total annotation effort thus amounts to approximately 170--250 person-hours.

%% file: sections/app_rubric_checklist.tex
\section{Details of the Hallucination Taxonomy}
\label{app:app_tax}

This section provides the full specification of our two-layer hallucination taxonomy.

\subsection{Layer 1: Substantive Legal Hallucination}
\label{app:tax_layer1}

Layer~1 captures errors in the \emph{content} of legal reasoning, organized along the canonical structure of legal analysis: \emph{source} $\rightarrow$ \emph{rule} $\rightarrow$ \emph{procedure} $\rightarrow$ \emph{application}. It comprises 4 mid-level categories and 19 fine-grained subcategories.

\subsubsection{L1.1 Authority Hallucination}
\label{app:tax_l11}

\paragraph{Scope.} All tasks requiring citation of authoritative legal sources (statutes, judicial interpretations, case numbers, official documents).

\begin{itemize}
\item \textbf{L1.1.1 Source Fabrication.} The model fabricates a non-existent statute, judicial interpretation, case number, or official document. \emph{Example:} ``Article~1300 of the Civil Code provides\ldots'' (the Civil Code has only 1260 articles).

\item \textbf{L1.1.2 Citation--Content Misapplication.} The cited source genuinely exists but is either inapplicable to the case at hand or its content is described inaccurately. \emph{Example:} ``Article~1062 of the Civil Code governs joint marital debts'' (it actually governs joint marital property).

\item \textbf{L1.1.3 Hierarchy Error.} The model confuses the hierarchical rank, territorial scope, or departmental authority of legal sources. \emph{Example:} Invoking a local regulation to override a superior statute; treating an administrative regulation as a ``law.''

\item \textbf{L1.1.4 Granularity Error.} The statute name and article number are both correct, but the specific \emph{paragraph}, \emph{item}, or \emph{sub-item} is wrong. \emph{Example:} Article~142, Paragraph~1 $\rightarrow$ incorrectly cited as Paragraph~2.
\end{itemize}




\subsubsection{L1.2 Doctrine Hallucination}
\label{app:tax_l12}

\paragraph{Scope.} Tasks examining substantive-law concepts, constituent elements, exceptions, legal consequences, and doctrinal positions.

\begin{itemize}
\item \textbf{L1.2.1 Conceptual Confusion.} The model misidentifies concept~$A$ as concept~$B$ at the categorical level, causing the entire reasoning chain to follow the wrong doctrinal branch. \emph{Example:} Apparent agency $\rightarrow$ unauthorized agency; guarantee $\rightarrow$ debt assumption; material misunderstanding $\rightarrow$ fraud.

\item \textbf{L1.2.2 Element Misstatement.} The correct legal concept is identified, but one or more of its constituent elements, formation conditions, or applicability thresholds are stated incorrectly. \emph{Example:} Correctly identifying apparent agency but claiming it requires ``fault'' (the actual standard is ``good faith + reasonable reliance'').

\item \textbf{L1.2.3 Exception Omission.} A legally decisive exception, defense, justification, or limiting condition is omitted. \emph{Example:} Analyzing contract validity without mentioning that ``unconscionability renders the contract voidable''; discussing breach liability without noting force majeure exemption. \emph{Note:} This subcategory is triggered only when the omitted exception is outcome-determinative (e.g., it distinguishes MCQ distractors or constitutes the core of an open-ended answer).

\item \textbf{L1.2.4 Consequence Error.} The legal consequence, form of liability, remedy, or penalty is stated incorrectly. \emph{Example:} Stating ``damages'' when the correct remedy is ``restitution''; concluding ``voidable'' when the contract is in fact ``void.''

\item \textbf{L1.2.5 Doctrinal Position Confusion.} The model conflates the prevailing view with minority positions, or inconsistently switches between doctrinal stances within the same analysis. \emph{Example:} Mixing subjective and objective theories of causation in criminal law; alternating between ``not yet formed'' and ``void'' to describe the same contract within one response.

\item \textbf{L1.2.6 Discretionary-Judgment Error.} Indeterminate legal concepts (e.g., ``relatively serious circumstances,'' ``unconscionability,'' ``sufficient to overturn the original judgment'') are applied outside their accepted boundaries. \emph{Example:} Misjudging the threshold for ``serious circumstances'' in sentencing; incorrectly applying the boundary of unconscionability in civil law.
\end{itemize}

\paragraph{Distinguishing L1.2.1 from L1.2.2.} The key diagnostic question is: ``Did the model take the wrong conceptual branch?'' If yes $\rightarrow$ L1.2.1. If the concept is correctly identified but individual elements are misstated $\rightarrow$ L1.2.2.

\subsubsection{L1.3 Procedural-Law Hallucination}
\label{app:tax_l13}

\paragraph{Scope.} Tasks involving civil procedure, criminal procedure, administrative litigation, or arbitration.

\begin{itemize}
\item \textbf{L1.3.1 Jurisdiction Error.} Errors in hierarchical, territorial, exclusive, or agreed jurisdiction. \emph{Example:} Assigning a case to a basic-level court when the amount in controversy exceeds its jurisdictional threshold.

\item \textbf{L1.3.2 Period Error.} Errors in statutes of limitation, peremptory periods, appeal deadlines, or filing periods. \emph{Example:} Stating the general civil limitation period as 2~years (correct: 3~years); stating the civil appeal period as 10~days (correct: 15~days).

\item \textbf{L1.3.3 Procedural-Step Error.} Errors in the sequence or requirements of procedural steps such as filing, acceptance, defense, evidence submission, preservation, or enforcement. \emph{Example:} Claiming the defense period is 30~days (correct: 15~days); asserting that property preservation requires prior filing of suit (pre-suit preservation is available).

\item \textbf{L1.3.4 Procedural-Outcome Error.} Errors in procedural dispositions---confusing \emph{dismissal of the action} (procedural defect) with \emph{dismissal of the claim} (substantive ruling), or misidentifying withdrawal or default judgment. \emph{Example:} Concluding ``dismiss the claim'' for what is actually a jurisdictional defect warranting ``dismiss the action.''

\item \textbf{L1.3.5 Appeal Error.} Errors in remedy paths including second-instance appeal, retrial, prosecutorial protest (\emph{kangsu}), administrative reconsideration, or enforcement objections. \emph{Example:} Advising direct litigation when administrative reconsideration is a mandatory prerequisite.
\end{itemize}

\subsubsection{L1.4 Application \& Subsumption Hallucination}
\label{app:tax_l14}

\paragraph{Scope.} All tasks requiring extraction of facts from a case narrative and mapping them onto legal rules.

\begin{itemize}
\item \textbf{L1.4.1 Fact Fabrication.} The model invents facts, amounts, dates, parties, or acts not present in the problem statement. \emph{Example:} The prompt contains no mention of ``divorce by agreement,'' yet the model states ``the parties divorced by agreement.''

\item \textbf{L1.4.2 Fact Omission.} A material fact is overlooked or a non-material fact is treated as decisive. \emph{Example:} Failing to note that ``the defendant is a state functionary,'' leading to omission of a bribery charge.

\item \textbf{L1.4.3 Element--Fact Mismatch.} The legal rule and its elements are stated correctly, but the facts are mapped to the wrong elements. \emph{Example:} The elements of apparent agency are correctly recited, but the model claims ``the counterparty was unaware'' satisfies ``good faith'' when the prompt explicitly states the counterparty had knowledge.

\item \textbf{L1.4.4 Party Confusion.} The identities, procedural positions, or rights/obligations of parties are mixed up. \emph{Example:} Reversing ``Party~A'' and ``Party~B''; treating an agent as the principal.
\end{itemize}

\paragraph{Distinguishing L1.4.3 from L1.2.2.} L1.2.2 addresses errors in stating the elements themselves; L1.4.3 addresses correct elements incorrectly matched to the case facts.

\subsection{Layer 2: Agent-Procedural Hallucination}
\label{app:tax_layer2}

Layer~2 targets failures along the agent's \emph{decision-making trajectory}, independent of substantive legal content. It comprises 3 mid-level categories and 9 fine-grained subcategories.

\subsubsection{L2.1 Planning \& Reasoning Hallucination}
\label{app:tax_l21}

\paragraph{Scope.} Reasoning subcategories can be detected within a single rollout; planning subcategories typically require an agent trajectory.

\paragraph{Distinction from L1.4.3.} L1.4.3 captures element-to-fact correspondence errors; L2.1 captures breakdowns in the logical chain itself (deduction, analogy, causation).

\begin{itemize}
\item \textbf{L2.1.1 Premature Closure.} The model locks onto a conclusion before sufficient information has been processed, and subsequent reasoning serves only to rationalize the premature conclusion. \emph{Example:} Determining the criminal charge before fully reading the case facts, then selectively citing statutes that support only that charge.

\item \textbf{L2.1.2 Syllogism Error.} The deductive chain (major premise [statute] $\rightarrow$ minor premise [facts] $\rightarrow$ conclusion) is broken or inverted. \emph{Example:} The statute requires ``intent,'' the facts establish only ``negligence,'' yet the model concludes intent is satisfied.

\item \textbf{L2.1.3 Self-Contradiction.} Conclusions, element selections, or legal characterizations within a single output or across steps are mutually inconsistent. \emph{Example:} First stating ``the contract is valid,'' then later stating ``the contract is voidable due to unconscionability''; step~5 contradicts the statutory basis cited in step~2.

\item \textbf{L2.1.4 Step Skip / Conflation.} Steps that require separate analysis are collapsed into one, or the model jumps from facts directly to a judgment without doctrinal grounding. \emph{Example:} A case analysis that directly states ``Judgment: the defendant bears full liability'' without identifying elements or citing statutes.

\item \textbf{L2.1.5 Out-of-Context Quoting.} A statutory provision is partially quoted as authority while contextual qualifications are suppressed. \emph{Example:} Quoting ``a loan contract is formed when the lender provides the loan'' while omitting the qualifying phrase ``between natural persons.''
\end{itemize}

\paragraph{Note on L2.1.1.} A writing style that states the conclusion first and then provides complete supporting analysis does \emph{not} constitute Premature Closure. L2.1.1 requires both: (a)~the model fails to genuinely consider alternative options or counterexamples after committing to an answer, and (b)~the reasoning trace shows no balanced analysis of competing candidates.

\subsubsection{L2.2 Memory Hallucination}
\label{app:tax_l22}

\paragraph{Scope.} Multi-turn dialogues or long agent trajectories only.

This category captures failures in maintaining consistency with information established earlier in the interaction. Specifically, it covers three manifestations: (1)~forgetting premises from prior turns, including the system prompt, the user's original request, or previously confirmed facts; (2)~misremembering key information such as party names, monetary amounts, dates, or the focal issues in dispute; and (3)~drifting away from the system prompt or the user's original question---e.g., the user asks about topic~$A$ but the model responds about topic~$B$.

\subsubsection{L2.3 Tool-Call \& Observation Hallucination}
\label{app:tax_l23}

\paragraph{Scope.} Trajectories involving tool invocations only. We distinguish two failure modes: errors in \emph{calling} tools versus errors in \emph{interpreting} tool outputs.

\begin{itemize}
\item \textbf{L2.3.1 Tool-Call Error.} The wrong tool is selected, parameters are incorrect, invocation order is wrong, a necessary call is omitted, or redundant calls are made. \emph{Example:} Using a statute-retrieval tool to search for case law; passing an incorrect article number to the retrieval API; using a generic search engine when a judicial-interpretation API is available.

\item \textbf{L2.3.2 Observation Misuse.} The tool returns correct content, but the model misreads or misinterprets the output. \emph{Example:} The retrieval tool returns five relevant statutes but the model only reads the first; a case date of 2018 is misread as 2008; the model relies on a summary snippet rather than the full text.
\end{itemize}

%% file: sections/app_judge.tex
\section{Judge}
\label{app:judge}

\subsection{Metrics}

For each rollout $r$, let $\hat Y(r)$ denote the predicted answer extracted from the model's free-form output, and let $Y(r)$ denote the gold answer. Answer correctness is defined as:
\begin{equation}
\mathrm{AC}(r) = \mathbb{I}[\hat{Y}(r) = Y(r)] \in \{0, 1\},
\label{eq:ac}
\end{equation}
the Substantive Cleanliness and the Procedural Cleanliness are defined as follows:
\begin{equation}
\mathrm{SC}(r) =\mathbb{I}(\sum_{s\in L1} h_{r,s}=0)\in\{0,1\},
\label{eq:sc}
\end{equation}
\begin{equation}
\mathrm{PC}(r) =\mathbb{I}(\sum_{s\in L2} h_{r,s}=0)\in\{0,1\}.
\label{eq:pc}
\end{equation}
We then define the Right Answer Wrong Reasoning (RAWR) rate. The RAWR rate for the above metric can be calculated as
\begin{equation}
RAWR-S =P(SC=0|AC=1),
\label{eq:sc}
\end{equation}
\begin{equation}
RAWR-P =P(PC=0|AC=1).
\label{eq:sc}
\end{equation}

\paragraph{Co-occurrence Matrix: Sections II--V} We also define the hallucination frequency for each subclass $s$ as follows:
\begin{equation}
P(s) \;=\; \frac{1}{N}\sum_{r=1}^{N} h_{r,s},
\end{equation}
represents the hit rate of subclass $s$ across all rollouts. Thus, the \textbf{Joint Halluciantion Frequency of a Pair of Subclasses} can be derived by

\begin{equation}
P(s,t) \;=\; \frac{1}{N}\sum_{r=1}^{N} h_{r,s}h_{r,t},
\end{equation}
which represents the fraction of rollouts hit by subclass $s$ and $t$ simultaneously. Note that $h_{r,s}h_{r,t}=1$ only when both indicators are 1.

Then, we derive the Lift metric:

\begin{equation}
\mathrm{Lift}(s,t) = \frac{P(s,t)}{ P(s)\,P(t)},
\end{equation}
by applying Bayes' Theorem, which can be further simplified as
\begin{equation}
\mathrm{Lift}(s,t) = \frac{P(s,t)/P(s)}{P(t)}=\frac{P(t|s)}{P(t)}.
\end{equation}

\paragraph{Interpretation}

Lift is a standard measure from association rule mining \citep{10.1145/170036.170072}: $\mathrm{Lift}(s,t) = 1$ indicates that $s$ and $t$ co-occur exactly as often as expected under independence; $\mathrm{Lift}(s,t) > 1$ indicates positive association, meaning a rollout exhibiting $s$ is more likely than chance to also exhibit $t$ (and vice versa); and $\mathrm{Lift}(s,t) < 1$ indicates negative association, meaning the two error types co-occur less often than independence would predict.




\subsection{Judge Robustness}
\label{app:judge_robustness}

To validate the viability of our evaluation, we engaged three legal experts, all of whom hold the Certificate of Legal Professional Qualification of the People's Republic of China, to label 2,391 rubric items and 1,395 valid judge decisions independently. For each item, we measured both the consistency between the LLM judge and human annotations and the inter-annotator agreement among the three experts. The results are summarized in Table~\ref{tab:judge_robustness}.

\begin{table*}[h]
\centering
\label{tab:judge_robustness}
\begin{tabular}{lcc}
\toprule
\textbf{Reliability} & \textbf{LLM--human consistency} & \textbf{Inter-annotator (three-expert) agreement} \\
\midrule
Rubric & 94.22\% & 91.00\% \\
Judge  & 86.90\% & 95.07\% \\
\bottomrule
\end{tabular}
\caption{Reliability of the rubric construction and judge scoring, measured by LLM--human consistency and inter-annotator (three-expert) agreement.}
\end{table*}

Taken together, these results demonstrate the robustness of our judge: the high inter-expert agreement confirms that the judgments are well-defined and stable across independent experts, while the judge--human consistency shows that our automated pipeline aligns closely with expert assessment.

%% file: sections/app_exprimental_details.tex
\section{Detailed Experimental Setup}
\label{app:app_exp_set}

\subsection{Hardware and Compute Environment}
All experiments run on a single Slurm cluster of DGX nodes, each
equipped with 8$\times$NVIDIA H800 80\,GB GPUs (224\,vCPU, $\sim$2\,TB
host RAM). Open-source backbones and the judge LLM are served by
vLLM~0.6.x in a Python~3.12 / CUDA~12.8 environment; closed-source
models are queried through an OpenAI-compatible reseller endpoint and
therefore consume no local GPUs.




\subsection{Generation Parameters}
\label{app:decoding}
All systems are evaluated under an identical inference configuration
with deterministic decoding: temperature $= 0.0$ (greedy), top\_p $= 0.8$,
top\_k $= 20$ , and max\_tokens $= 4{,}096$ per
LLM call.

\subsection{Agentic Workflow Parameters}
\label{app:workflows}

The three orchestrations are configured as follows.
\begin{itemize}\itemsep0pt
\item \textbf{ReAct}: \texttt{max\_steps} $= 8$. Each step emits either
  \texttt{<action>}\,$\to$\,\texttt{Observation} or
  \texttt{<final\_answer>}; budget exhaustion triggers a force-finalise
  step that composes the answer from the existing trace.
\item \textbf{Plan-and-Execute}: \texttt{max\_steps} $= 10$. A planner
  first emits a typed JSON plan; each plan step is executed through the
  same tool catalogue as ReAct; on budget exhaustion the agent likewise
  produces a force-finalised answer.
\item \textbf{LawThinker}: a two-turn scenario between a rule-based
  Questioner and a LawThinker trainee (\texttt{Agent.Trainee.LC\_Generic}),
  with an optional \texttt{deep\_analysis} sub-stage that runs a second
  pass over the answer; \texttt{LAWTHINKER\_DISABLE\_THINKING=1}
  suppresses backbone-native chain-of-thought tags so the only reasoning
  trace observable to the judge is the explicit dialog history.
\end{itemize}

\subsection{Shared Tool Suite}
\label{app:tool_suite}

ReAct, Plan-and-Execute, and LawThinker share an identical legal tool
suite. All tools are exposed verbatim in the agent system prompts; the
exact same string is replayed to the judge LLM (\texttt{tool\_specs.py})
to enable the L2.3 tool-misuse sub-class to be evaluated byte-for-byte
against the spec the model actually saw. The catalogue is partitioned
into three families:

\paragraph{(a) Knowledge-exploration tools (7).}
Used to surface candidate authorities and analogous artefacts when the
agent has only a natural-language question:
\begin{itemize}\itemsep0pt
\item \texttt{law\_retrieval(query, topk)} -- top-$k$ statute retrieval
  over a Chinese statute index (BGE-M3 dense embeddings); returns the
  full text of each candidate article.
\item \texttt{law\_recommendation(law)} -- given an explicit article
  reference (e.g.\ ``
Article 201 of the Criminal Law''), return statutes that frequently
  co-cite with it.
\item \texttt{charge\_expansion(charges)} -- expand a list of criminal
  charges into closely related charges from the same chapter or
  Supreme-Court interpretation.
\item \texttt{case\_retrieval(type, query)} -- analogous-case retrieval
  in the civil or criminal sub-corpus.
\item \texttt{template\_retrieval(template\_type)} -- fetch the official
  template for a procedural document (e.g.\ statement of complaint,
  defence).
\item \texttt{plan\_generation(document\_type)} -- emit a section
  skeleton for the requested document.
\item \texttt{procedure\_retrieval(court\_type, stage)} -- retrieve the
  civil (5-stage) or criminal (3-stage) court procedure; only relevant
  in the moot-court setting and rarely fired on \textsc{\ourbench}.
\end{itemize}

\paragraph{(b) Verification tools (6).}
Used after a candidate authority has been picked, to confirm
applicability and consistency:
\begin{itemize}\itemsep0pt
\item \texttt{law\_check(law\_name)} -- exact-text lookup for a given
  article reference; the canonical way to verify a citation.
\item \texttt{fact\_law\_relevance\_check(fact, law)} -- decide whether
  a given article applies to a given fact pattern.
\item \texttt{crime\_law\_consistency\_check(crime, law)} -- verify that
  a charge and a cited article actually correspond under criminal law.
\item \texttt{document\_format\_check(document\_type, document)} --
  format-validate a drafted procedural document.
\item \texttt{law\_query\_rewrite(query, context)} -- rewrite a vague
  query into a statute-grounded one before re-issuing
  \texttt{law\_retrieval}.
\item \texttt{procedure\_check(court\_type)} -- check completeness of a
  proposed civil/criminal procedure trajectory.
\end{itemize}

\paragraph{(c) Web fallback (1).}
\begin{itemize}\itemsep0pt
\item \texttt{web\_search(query, summarizer\_version)} -- public-web
  retrieval via a Bright Data SERP API, followed by Jina-Reader page
  fetch and a question-aware summariser (default
  \texttt{summarizer\_version=original}, max 5 hits per query). Reserved
  for queries that local retrieval cannot satisfy (latest policies,
  local regulations, foreign law, procedural ``how-to'' questions).
\end{itemize}

A short \textbf{tool-use discipline} is appended to every system prompt
(``rules of engagement''): identical \texttt{(name, arguments)} tuples
may not be re-issued; \texttt{law\_retrieval}/\texttt{law\_check} are
allowed at most two misses per dispute before the agent must escalate to
\texttt{web\_search} or close; one good \texttt{web\_search} hit must be
followed by \texttt{<final\_answer>}; common-sense or
elementary-doctrine questions should close immediately without invoking
any tool. \textbf{Memory tools}
(\texttt{memory\_store}/\texttt{memory\_fetch}) are deliberately removed
from the catalogue for the single-rollout benchmark setting; their
implementations remain in the runtime for multi-turn scenarios.

ReAct and Plan-and-Execute expose all 14 tools in (a)+(b)+(c) at once.
LawThinker exposes the (a) family during the initial response phase and
the (b) family during the deep-analysis verification phase, mirroring
its two-stage protocol.

%% file: sections/app_detailed_analysis.tex
\section{Supplementary Detailed Analysis}
\label{app:detailed_analysis}

\subsection{Framework Specialization (Per-Group Profile)}
\label{app:framework_spec}

Figure~\ref{fig:app_framework_spec} reports the within-group any-hit rate
HF$_g$ for each of the four agent frameworks across the seven
hallucination groups.  The view complements main paper finding~1
(\emph{Framework-Dependent Hallucination Profiles}) by exposing the
specific group on which each framework concentrates errors:
LawThinker / Plan-and-Execute / ReAct sit within $\pm 2$\,pp of each
other on every L1 group, while LRAS uniquely lifts L1.2 / L1.4
(Doctrine, Application) by $7$--$8$\,pp---an expected side-effect of
its self-RL training distribution---but lowers L2.3 (Tool/Obs) by the
largest margin.  This directly motivates the simplicity-vs.-diversity
discussion in the main paper's tool-tradeoff analysis.


\subsection{Per-Task and Per-Domain Difficulty}
\label{app:task_domain}

Table~\ref{tab:task_difficulty} reports the full HF/HD numbers for the
six task types of \textsc{\ourbench} (Figure~\ref{fig:task_difficulty}
visualises only HF and annotates HD).  Figure~\ref{fig:app_legal_category}
reports the same metrics for the top 12 legal categories, sorted by
HF$_{L1}$ descending.

\input{table/table_task_difficulty}

%% file: table/table_task_difficulty.tex

\begin{table}[t]
\centering\small
\setlength{\tabcolsep}{5pt}
\renewcommand{\arraystretch}{1.10}
\begin{tabular}{lrr}
\toprule
Task type & HD$_{L1}$ & HD$_{L2}$ \\
\midrule
Adjudication Analysis & 0.492 & 0.403 \\
Case Analysis         & 0.293 & 0.304 \\
Legal Reasoning       & 0.240 & 0.259 \\
Legal Consultation    & 0.213 & 0.182 \\
Legal Knowledge QA    & 0.197 & 0.208 \\
Judgement Prediction  & 0.215 & 0.261 \\
\bottomrule
\end{tabular}
\caption{Per-task-type hallucination profile on \textsc{\ourbench}, sorted by HD$_{L1}$ descending.}
\label{tab:task_difficulty}
\end{table}

%% file: sections/app_prompts_template.tex
\section{Prompt Templates}
\label{app:app_prompts}

\subsection{Rollout Prompt}

In this section, we provide the rollout-stage prompts for LawThinker, LRAS, Plan-and-Execute, and ReAct.

\input{sections/prompts/rollout_lt_1}

\input{sections/prompts/rollout_lt_2}

\input{sections/prompts/rollout_lras}

\input{sections/prompts/rollout_react}

\input{sections/prompts/rollout_pe_plan}

\input{sections/prompts/rollout_plan_execute}

\subsection{Data Filter Prompt}

In this section, we provide the prompts used to filter open-ended questions of different types during data curation.

\input{sections/prompts/filter_consultation}

\input{sections/prompts/filter_essay}

\input{sections/prompts/filter_judgment}

\subsection{Rubric checklist Annotation Prompt}

\input{sections/prompts/rubric_pretag}

\subsection{Judge Prompt}

This section presents the seven judge prompts used in our evaluation pipeline, each targeting one Layer-1 or Layer-2 hallucination group: L1.1 Authority, L1.2 Doctrine, L1.3 Procedural, L1.4 Application, L2.1 Planning \& Reasoning, L2.2 Memory, and L2.3 Tool \& Observation.

\input{sections/prompts/judge_L1_1_authority}
\input{sections/prompts/judge_L1_2_doctrine}

\input{sections/prompts/judge_L1_3_procedural}
\input{sections/prompts/judge_L1_4_application}
\input{sections/prompts/judge_L2_1_planning_reasoning}
\input{sections/prompts/judge_L2_2_memory}
\input{sections/prompts/judge_L2_3_tool_observation}

%% file: sections/prompts/rollout_lt_1.tex

\begin{promptbox}{LawThinker Instruction}
You are a legal-reasoning assistant that may call domain-specific legal tools whenever necessary.

Available tools
memory_fetch - retrieve stored knowledge or context.
law_retrieval - retrieve the top-k most relevant statutes given a natural-language query.
law_recommendation - return statutes similar to the one provided.
charge_expansion - expand a list of charges with related ones.
case_retrieval - retrieve similar civil/criminal cases.
template_retrieval - fetch a document template (e.g. complaint document, defence document).
writing_plan_generation - generate a writing plan for the given document type.
procedure_retrieval - retrieve the civil/criminal court procedure (stage=0 for the full procedure).

Tool-calling format
Whenever a tool is needed, output
<tool_call>{"name": ..., "arguments": ...}</tool_call>.
The system will respond with
<tool_call_result> ... </tool_call_result>.

Example: examples
\end{promptbox}

%% file: sections/prompts/rollout_lt_2.tex

\begin{promptbox}{DeepVerifier Prompt}
You are a deep-analysis legal assistant. Given (i) the user's last query or response, (ii) the current reasoning trace, and (iii) the result of the exploration step, decide whether the retrieved information is correct and relevant; decide whether more exploration is needed; and store any key facts or statutes.

Tool principles
1. Choose the appropriate checking tool to verify the accuracy and relevance of retrieved knowledge.
2. Summarize key information and store it with memory_store.

Available tools
memory_fetch - retrieve stored knowledge / context
memory_store - store key knowledge / context
law_article_check - verify law article content
fact_law_relevance_check - check law applicability to facts
charge_law_consistency_check - check charge-law consistency
search_query_rewrite - rewrite law retrieval queries
document_format_check - check document format
procedure_check - Check procedural compliance

Tool-calling format
<tool_call>{"name": ..., "arguments": ...}</tool_call>
<tool_call_result> ... </tool_call_result>.

Example: examples
\end{promptbox}

%% file: sections/prompts/rollout_lras.tex

\begin{promptbox}{LRAS Prompt}
You are a professional legal assistant skilled in answering legal questions accurately through in-depth research.

Workflow:
1. First conduct reasoning and analysis within the  tags. If you determine that existing knowledge is insufficient to answer accurately, perform a search.
2. To find legal authorities, use <search>query</search> to call the search tool; results will be returned within <information></information>.
3. You may perform multiple searches until sufficient information is obtained.
4. Finally, provide your answer within . Answer multiple-choice questions concisely; provide detailed answers for non-multiple-choice questions.

Question:
<<QUESTION>>
\end{promptbox}

%% file: sections/prompts/rollout_react.tex

\begin{promptbox}{ReAct Prompt}
[ReAct system]
You are a legal assistant following the ReAct (Thought-Action-Observation) paradigm. Your goal is to call tools when necessary and deliver accurate, interpretable legal responses.

### Tool List
- law_retrieval: Legal provision retrieval. Return the top-k most relevant provisions given a natural language query. Parameters: {"query": "str", "topk": "int"}
  Usage Guidelines:
  (1) Priority: Use law_check to obtain full text if the exact article number is known (e.g. Criminal Law Article 397); use law_retrieval for natural language descriptions only.
  (2) Query Writing: Focus on "crime/legal system name + key constitutive elements" or "specific article number", avoid complete questions; topk is set to 3-5 by default.
  (3) Poor Retrieval Results: Retry once with revised keywords. If no valid results are obtained, switch to web_search or make a conclusion based on existing information. Do not repeatedly retry the same query.
  (4) Coverage: Current effective statutes, judicial interpretations and departmental rules of Chinese mainland. Low hit rate for local regulations, newly released policies, international conventions and foreign laws.
- law_recommendation: Similar provision recommendation. Return similar provisions given a specified legal article (e.g. Criminal Law Article 201). Parameters: {"law": "str"}
- charge_expansion: Related charge expansion. Return similar charges based on the given charge list. Parameters: {"charges": "List[str]"}
- case_retrieval: Similar case retrieval. Return analogous cases given case type (civil or criminal) and case information. Parameters: {"type": "str(Civil Case|Criminal Case)", "query": "str"}
- template_retrieval: Document template retrieval. Obtain templates for specified legal documents (e.g. complaint, defense statement). Parameters: {"template_type": "str(Complaint|Defense Statement)"}
- plan_generation: Writing plan generation. Create a writing plan according to the document type. Parameters: {"document_type": "str(Complaint|Defense Statement)"}
- procedure_retrieval: Court procedure retrieval. Only applicable to moot court scenarios. Stage stands for proceeding phase; set stage=0 for the full procedure. Civil proceedings contain 5 phases and criminal proceedings contain 3 phases. Parameters: {"court_type": "str(Civil Court|Criminal Court)", "stage": "int(0-4|0-2)"}
- law_check: Legal provision verification. Return the full text of a specified legal article (e.g. Criminal Law Article 201). Parameters: {"law_name": "str"}
  Usage Guideline: Prioritize law_check for higher accuracy when the exact article number is available.
- fact_law_relevance_check: Fact-provision relevance verification. Determine whether a given legal article (e.g. Criminal Law Article 201) applies to the case facts. Parameters: {"fact": "str", "law": "str"}
- crime_law_consistency_check: Charge-provision matching verification. Verify whether the given charge matches the specified criminal law article (e.g. Criminal Law Article 201). Parameters: {"crime": "str", "law": "str"}
- document_format_check: Document format inspection. Check the format of legal documents. Submit complete document content without omission. Parameters: {"document_type": "str(Complaint|Defense Statement)", "document": "str"}
- law_query_rewrite: Query rewriting. Revise the original query combined with case background to make it clearer for retrieval. Parameters: {"query": "str", "context": "str"}
- procedure_check: Proceeding inspection. Verify the completeness of civil or criminal court procedures. Parameters: {"court_type": "str(Civil Court|Criminal Court)"}
- web_search: Web search. Retrieve legal information from public internet and generate key summaries related to the user's question.
  Application Scenarios:
  (1) No valid results after one keyword revision via law_retrieval / law_check;
  (2) Inquiries involving newly released policies, local regulations, details of typical cases, international conventions or foreign laws;
  (3) Consultation on practical procedures, required materials and service channels.
  Rules:
  (1) Compile summaries and deliver the final answer once relevant results are returned. Do not resubmit the same query.
  (2) Explicit rules such as "The insured shall..." or "In accordance with Article XX of XX Law" in summaries are deemed valid results. You may refine the query at most once before concluding the task.
  (3) Run web_search no more than 2 times for a single disputed point.
  Parameters: {"query": "str (Chinese keywords, within 25 characters recommended)", "summarizer_version": "str(original|v1) (optional, original by default)"}

### Tool Usage Rules (Must be strictly followed)
1. No Duplicate Calls: Do not call the same tool with identical parameters twice. For retries, adjust keywords, topk value or switch to another tool.
2. Upgrade Path: If law_retrieval / law_check fails to return valid results for over 2 times regarding one disputed point, immediately switch to web_search or draw a conclusion. If web_search returns relevant summaries, stop searching and output the final answer directly.
3. Prioritize Conclusion: Give the final answer directly if the question can be judged based on general knowledge or basic legal principles. For multiple-choice questions with clear legal rationales for all options, stop tool calls immediately.
4. Tool Selection: Use law_check if the article number is confirmed; use law_retrieval for natural language descriptions; use web_search for newly released policies, local regulations, service procedures and foreign laws.

### Examples (For format reference only)
Example 1 (Tool Call):
Thought: The user asks about constitutive elements of fraud, I will retrieve relevant criminal law provisions first.
<action>{"name":"law_retrieval","arguments":{"query":"fraud constitutive elements","topk":3}}</action>

Example 2 (Final Answer):
Thought: Criminal Law Article 266 has been retrieved, I can answer the question based on the facts.
<final_answer>Conclusion: This act constitutes fraud. Basis: Criminal Law Article 266 states that ...</final_answer>

### Output Rules (Must be strictly followed)
1. Choose one format for each round:
   - Tool Call:
     Thought: ...
     <action>{"name":"Tool Name","arguments":{...}}</action>
   - Final Answer:
     Thought: ...
     <final_answer>Final answer for the user</final_answer>
2. The tool name in <action> must be selected from the tool list, and arguments must be valid JSON.
3. Output only one <action> or one <final_answer> per round. Do not include both.
4. Do not add irrelevant labels such as Observation or Step numbers.

[ReAct user turn template]
### Current Question
<<QUESTION>>

### Current Reasoning Trace
<<SCRATCHPAD>>

Continue with the next step. Maximum 8 rounds allowed.
\end{promptbox}

%% file: sections/prompts/rollout_pe_plan.tex

\begin{promptbox}{Plan-and-Execute Prompt--Planner system}
You are a legal task planner. Generate executable plans in JSON format for legal questions raised by users.

Output Rules (Must be strictly followed):
1. Output a single JSON object only, with no explanations or Markdown code blocks.
2. The top-level JSON must contain the field "plan" whose value is a string array.
3. The plan shall include at least 2 steps. There is no upper limit, while 3 to 6 actionable and progressive steps are recommended.
4. Each step must be specific and executable (including which tool to call and which dispute points to focus on). Vague expressions such as "think about it" or "make an analysis" are prohibited.

Example (Multiple Choice Question):
{
  "plan": [
    "Sort out key facts of the question and identify disputed points (e.g. conduct nature, subjective state, role identity)",
    "Verify relevant legal provisions and constitutive elements for each option respectively",
    "Determine the only correct answer and briefly explain why other options are excluded"
  ]
}

Example (Consultation Question):
{
  "plan": [
    "Extract the user's facts and demands, and identify the relevant legal field",
    "Retrieve applicable laws and judicial interpretations; switch to web_search if no results are obtained locally",
    "Draw conclusions combined with facts, deliver risk reminders and feasible suggestions"
  ]
}
\end{promptbox}

%% file: sections/prompts/rollout_plan_execute.tex

\begin{promptbox}{Plan-and-Execute Prompt--Excute System}
You are a Plan-and-Execute legal assistant. Carry out tasks in accordance with the plan, call tools when necessary, and finally deliver explainable conclusions.

### Tool List
- law_retrieval: Legal provision retrieval. Return the top-k most relevant provisions given a natural language query. Parameters: {"query": "str", "topk": "int"}
  Usage Guidelines:
  (1) Priority: Use law_check to obtain full text if the exact article number is known (e.g. Criminal Law Article 397); use law_retrieval for natural language descriptions only.
  (2) Query Writing: Focus on "crime/legal system name + key constitutive elements" or "specific article number", avoid using complete questions; topk is set to 3-5 by default.
  (3) Poor Retrieval Results: Retry once with revised keywords. If no valid results are obtained, switch to web_search or make a conclusion based on existing information. Do not repeatedly retry the same query.
  (4) Coverage: Current effective statutes, judicial interpretations and departmental rules of Chinese mainland. Low hit rate for local regulations, newly released policies, international conventions and foreign laws.
- law_recommendation: Similar provision recommendation. Return similar provisions given a specified legal article (e.g. Criminal Law Article 201). Parameters: {"law": "str"}
- charge_expansion: Related charge expansion. Return similar charges based on the given charge list. Parameters: {"charges": "List[str]"}
- case_retrieval: Similar case retrieval. Return analogous cases given case type (civil or criminal) and case information. Parameters: {"type": "str(Civil Case|Criminal Case)", "query": "str"}
- template_retrieval: Document template retrieval. Obtain templates for specified legal documents (e.g. complaint, defense statement). Parameters: {"template_type": "str(Complaint|Defense Statement)"}
- plan_generation: Writing plan generation. Create a writing plan according to the document type. Parameters: {"document_type": "str(Complaint|Defense Statement)"}
- procedure_retrieval: Court procedure retrieval. Only applicable to moot court scenarios. Stage stands for proceeding phase; set stage=0 for the full procedure. Civil proceedings contain 5 phases and criminal proceedings contain 3 phases. Parameters: {"court_type": "str(Civil Court|Criminal Court)", "stage": "int(0-4|0-2)"}
- law_check: Legal provision verification. Return the full text of a specified legal article (e.g. Criminal Law Article 201). Parameters: {"law_name": "str"}
  Usage Guideline: Prioritize law_check for higher accuracy when the exact article number is available.
- fact_law_relevance_check: Fact-provision relevance verification. Determine whether a given legal article (e.g. Criminal Law Article 201) applies to the case facts. Parameters: {"fact": "str", "law": "str"}
- crime_law_consistency_check: Charge-provision matching verification. Verify whether the given charge matches the specified criminal law article (e.g. Criminal Law Article 201). Parameters: {"crime": "str", "law": "str"}
- document_format_check: Document format inspection. Check the format of legal documents. Submit complete document content without omission. Parameters: {"document_type": "str(Complaint|Defense Statement)", "document": "str"}
- law_query_rewrite: Query rewriting. Revise the original query combined with case background to make it clearer for retrieval. Parameters: {"query": "str", "context": "str"}
- procedure_check: Proceeding inspection. Verify the completeness of civil or criminal court procedures. Parameters: {"court_type": "str(Civil Court|Criminal Court)"}
- web_search: Web search. Retrieve legal information from public internet and generate key summaries related to the user's question.
  Application Scenarios:
  (1) No valid results after one keyword revision via law_retrieval / law_check;
  (2) Inquiries involving newly released policies, local regulations, details of typical cases, international conventions or foreign laws;
  (3) Consultation on practical procedures, required materials and service channels.
  Rules:
  (1) Compile summaries and deliver the final answer once relevant results are returned. Do not resubmit the same query.
  (2) Explicit rules such as "The insured shall..." or "In accordance with Article XX of XX Law" in summaries are deemed valid results. You may refine the query at most once before concluding the task.
  (3) Run web_search no more than 2 times for a single disputed point.
  Parameters: {"query": "str (Chinese keywords, within 25 characters recommended)", "summarizer_version": "str(original|v1) (optional, original by default)"}

### Tool Usage Rules (Must be strictly followed)
1. No Duplicate Calls: Do not call the same tool with identical parameters twice. For retries, adjust keywords, topk value or switch to another tool.
2. Upgrade Path: If law_retrieval / law_check fails to return valid results for over 2 times regarding one disputed point, immediately switch to web_search or draw a conclusion. If web_search returns relevant summaries, stop searching and output the final answer directly.
3. Prioritize Conclusion: Give the final answer directly if the question can be judged based on general knowledge or basic legal principles. For multiple-choice questions with clear legal rationales for all options, stop tool calls immediately.
4. Tool Selection: Use law_check if the article number is confirmed; use law_retrieval for natural language descriptions; use web_search for newly released policies, local regulations, service procedures and foreign laws.

### Output Rules (Must be strictly followed)
1. Choose one format for each round:
   - Tool Call:
     Thought: <You may state "Now proceed with Step K">
     <action>{"name":"Tool Name","arguments":{...}}</action>
   - Final Answer:
     Thought: ...
     <final_answer>Final answer for the user</final_answer>
2. The tool name in <action> must be selected from the tool list, and arguments must be valid JSON.
3. Output only one <action> or one <final_answer> per round. Do not include both.
4. Focus on steps marked [doing] in the plan status. Explain the completion of current steps in Thought and proceed to the next step implicitly.
5. Do not add irrelevant labels such as Observation or Step numbers.

[Executor user turn template]
### Current Question
<<QUESTION>>

### Plan Status ([done] Completed / [doing] In Progress / [todo] Pending)
<<PLAN_STATUS>>

### Execution Progress (Including tool calls and original results)
<<PROGRESS>>

Continue with the next step. Maximum 10 rounds allowed.
\end{promptbox}

%% file: sections/prompts/filter_consultation.tex

\begin{promptbox}{Data Filter: LawBench 3-8 (Consultation)}
You are a strict data quality reviewer examining a sample of legal consultation Q\&A.

\textbf{Task Background}
- Question: Colloquial legal inquiries from general users, usually with redundant content, typos, missing subjects, mixed regional information and other issues.
- Answer: A standard answer consists of two parts:
  (a) A direct response to the user's question (marked with "Answer: ..." or equivalent expressions);
  (b) Citation of specific laws and regulations (marked with "Legal Basis: ..." or equivalent expressions).

\textbf{Key Principles (Must be strictly followed)}
1. You are neither required nor allowed to judge the accuracy of legal provisions cited, correctness of legal conclusions or rationality of judgments.
2. You shall score solely based on observable textual features, including structure, relevance, completeness and citation presence. Only judge existence rather than correctness.

\textbf{Scoring Criteria} (Score each item with an integer from 1 to 5)

1. Question Clarity
   - 5: The user's core demand is clearly identifiable.
   - 3: The general meaning is understandable, with redundant content or awkward expressions.
   - 1: The question is confusing with missing subjects or key information, and the inquiry cannot be figured out.

2. Question-Answer Relevance
   - 5: The answer clearly addresses the core demand and stays on topic.
   - 3: The answer is partially relevant but deviates or barely relates to the question.
   - 1: The answer is almost irrelevant or completely off-topic.

3. Answer Structural Compliance (Legal Structure Evaluation)
   - 5: The answer clearly contains a two-part structure: direct response plus legal basis. Wording may vary, but the two sections are distinguishable.
   - 3: Only one part is present (either a conclusion without legal basis, or listed provisions without a response).
   - 1: Neither part exists; the content is empty remarks or rhetorical questions.

4. Legal Citation Format (Only judge existence, \textbf{not} accuracy)
   - 5: At least one standard citation of specific laws or regulations appears (e.g., \textit{Article X of XX Law}, \textit{According to YY Rules}, \textit{Contract Part of the Civil Code}).
   - 3: Only general legal names are mentioned without specific provisions (e.g., merely stating "in accordance with civil law").
   - 1: No legal citation of any form is present.
   - Note: Do not verify the authenticity or correctness of citations; only check whether citation formats appear in the text.

5. Answer Completeness
   - 5: The answer is complete with no obvious truncation.
   - 3: Slightly incomplete but the main idea is intelligible.
   - 1: The answer ends abruptly with unfinished sentences or ellipses replacing key discussions.

6. Internal Consistency of Answer
   - 5: Statements in different parts are consistent and free of contradictions.
   - 3: Minor wording conflicts exist without affecting the overall conclusion.
   - 1: Obvious contradictions on the same conclusion appear across sections.

7. Substantiality of Answer
   - 5: The answer contains detailed reasoning and substantial content.
   - 3: The content is brief but delivers basic information.
   - 1: The answer is overly short, consists of empty remarks, repeats the question largely, or only gives a conclusion without any analysis.

\textbf{Output Format}
Output strictly in the following JSON format. Do not add any explanations, prefixes, suffixes or reasoning content.

```json
{
  "Question Clarity": <1-5>,
  "Question-Answer Relevance": <1-5>,
  "Answer Structural Compliance": <1-5>,
  "Legal Citation Format": <1-5>,
  "Answer Completeness": <1-5>,
  "Internal Consistency of Answer": <1-5>,
  "Substantiality of Answer": <1-5>,
  "Suggestion": "<Retain | Reject>",
  "Reason": "<Brief explanation within 50 words>"
}
\textbf{Judgment Reference (Comprehensive judgment based on scores above)}
Retain: All scores 
>=4
 and no score equals 1.
Reject: Average score ranges from 3.0 to 3.9, or any single item scores 2.
Reject: Average score 
<3.0
, or any single item scores 1.
\textbf{Question (User Inquiry)}
__QUESTION__
\textbf{Answer (Lawyer / AI Response)}
__ANSWER__
Please output the JSON scores strictly in the above format without any extra text outside the markdown code block:
\end{promptbox}

%% file: sections/prompts/filter_essay.tex
\begin{promptbox}{Data Filter: LexEval 5-4}
You are a rigorous data quality reviewer assessing samples of subjective questions for the National Judicial Examination.

Task Background
- Question: A case description followed by one or more specific questions (marked as "Q: 1) ...  2) ..."), which usually contains multiple sub-questions.
- Answer: Legal analysis addressing the raised questions. \textbf{No restrictions on answer formats}, including "Conclusion + Reasons", comparison of multiple viewpoints, step-by-step analysis combined with case facts, or citation of relevant precedents and legal theories. Provide responses separately for each sub-question if there are several.

Key Principles (Must be strictly followed)
1. You are neither required nor allowed to judge the accuracy of cited legal provisions, correctness of legal conclusions or reasonableness of judgments.
2. Score exclusively based on observable textual features such as structure, relevance, completeness and citation forms. Judge only the presence of elements rather than their correctness.

Scoring Criteria (Assign an integer score from 1 to 5 for each item)

1. Question Clarity
   - 5: The case facts and questions are clearly stated, with clear examination focuses.
   - 3: Readable with slightly awkward expressions in parts.
   - 1: The text is confusing and the questions are incomprehensible.

2. Completeness of Case Elements (Legal structure assessment)
   - 5: The case contains all key elements for answering, including parties, conducts, time and disputes.
   - 3: Basic elements are provided with minor omissions.
   - 1: Severe lack of key elements, leaving no factual basis for answering.

3. Question Coverage
   - 5: All sub-questions are responded to correspondingly.
   - 3: Most sub-questions are covered with one omission.
   - 1: Only one sub-question is answered while others are skipped entirely.
   - Note: Only check if responses exist, not whether the answers are correct.

4. Argumentation Quality (Legal structure assessment, no fixed format required)
   - 5: Each question is supported by identifiable legal analysis, containing clear positions/conclusions and corresponding arguments. Any valid structure is acceptable as long as both position and reasoning are included.
   - 3: Merely a position with weak arguments, or plain reasoning without clear conclusions.
   - 1: No recognizable legal analysis, such as repeating the question, listing provisions without analysis, or giving a conclusion with no reasons.
   - Note: Do not evaluate the validity of analysis; only check for the coexistence of positions and arguments.

5. Legal Citation Format (Only judge existence, not accuracy)
   - 5: At least one specific citation of laws or regulations is included.
   - 3: Only general legal names are mentioned without specific clauses.
   - 1: No legal citations of any kind are present.

6. Answer Completeness
   - 5: The answer is fully structured with no truncation.
   - 3: Slightly truncated but the main idea remains intact.
   - 1: Ends abruptly or uses long ellipses to replace key arguments.

7. Internal Consistency
   - 5: No contradictions between answers to different sub-questions, or between conclusions and supporting reasons.
   - 3: Minor conflicts that do not affect the main point.
   - 1: Obvious contradictions on the same conclusion.

8. Substantiality of Answer
   - 5: Each sub-question is elaborated with detailed legal arguments.
   - 3: Brief arguments with key points included.
   - 1: Vague arguments or a single-sentence conclusion without analysis.

Output Format
Strictly follow the JSON format below. Do not add any explanations, prefixes, suffixes or reasoning content.

```json
{
  "Question Clarity": <1-5>,
  "Completeness of Case Elements": <1-5>,
  "Question Coverage": <1-5>,
  "Argumentation Quality": <1-5>,
  "Legal Citation Format": <1-5>,
  "Answer Completeness": <1-5>,
  "Internal Consistency": <1-5>,
  "Substantiality of Answer": <1-5>,
  "Suggestion": "<Retain | Reject>",
  "Reason": "<Brief explanation within 50 words>"
}
Judgment Rules
Retain: All scores \(\ge 4\) and no score equals 1.
Reject: Average score ranges from 3.0 to 3.9, or any item scores 2.
Reject: Average score \(< 3.0\), or any item scores 1.
Question (Case & Questions)
__QUESTION__Answer (Standard Answer)
__ANSWER__Please output the JSON scores strictly in the above format without any extra text outside the markdown code block:
\end{promptbox}

%% file: sections/prompts/filter_judgment.tex

\begin{promptbox}[breakable]{Data Filter: LexEval 5-2}
You are a rigorous data quality reviewer examining samples of judgment analysis generation.

\textbf{Task Background}
- Question: Factual findings stated by the court. Ideally, it shall include parties, cause of action/legal relationship, contract or event process, and focus of dispute.
- Answer: Legal analysis and judgment document. The standard structure consists of two parts:
  (a) Argumentation section starting with "The court holds..." for legal analysis on disputed facts;
  (b) Clear judgment rulings (e.g., "Judgment as follows: ..." plus specific verdicts).

\textbf{Key Principles (Must be strictly followed)}
1. You are neither required nor allowed to judge the accuracy of cited legal provisions, correctness of legal conclusions or reasonableness of judgments.
2. Score solely based on observable textual features including structure, relevance, completeness and citation forms. Only judge the presence of elements rather than their correctness.

\textbf{Scoring Criteria} (Assign an integer score from 1 to 5 for each item)

1. Case Clarity
   - 5: Facts are stated coherently and clearly with good readability.
   - 3: Readable but slightly disorganized.
   - 1: Confusing and fragmented text, barely intelligible.

2. Completeness of Case Elements (Legal structure assessment)
   - 5: The question contains all three key elements: parties, cause of action/legal relationship, and disputed facts.
   - 3: One key element is missing.
   - 1: Multiple key elements are missing, and the trial focus cannot be identified.
   - Note: Do not assess rationality or authenticity of elements; only check their presence.

3. Question-Answer Relevance
   - 5: The judgment analysis clearly centers on the disputed facts in the given case.
   - 3: Partially relevant with off-topic content.
   - 1: The answer is almost irrelevant or completely off-topic.

4. Judgment Structural Compliance (Legal structure assessment)
   - 5: The answer contains both a recognizable argumentation section (marked by "The court holds" or equivalent expressions) and explicit judgment rulings.
   - 3: Only the argumentation section or only the judgment rulings are provided.
   - 1: Neither identifiable argumentation nor judgment rulings are included.

5. Legal Citation Format (Only judge existence, \textbf{not} accuracy)
   - 5: At least one specific citation of laws or regulations appears (e.g., Article X of the Civil Procedure Law).
   - 3: Only general legal names are mentioned without specific clauses.
   - 1: No legal citations of any form are present.

6. Answer Completeness
   - 5: The content is fully organized with complete and definite judgment rulings.
   - 3: Judgment rulings exist but are not fully stated.
   - 1: The text ends abruptly, with missing rulings or unfinished sentences.

7. Internal Consistency
   - 5: The viewpoints in the argumentation section are consistent with the final judgment rulings.
   - 3: Minor discrepancies without actual contradictions.
   - 1: Obvious contradictions between argumentation and rulings.
   - Note: Do not judge legal validity; only check textual consistency.

8. Substantiality of Answer
   - 5: The argumentation includes detailed factual analysis and legal reasoning.
   - 3: Concise argumentation with key points covered.
   - 1: The argumentation consists merely of formal remarks with no substantial analysis.

\textbf{Output Format}
Strictly follow the JSON format below. Do not add any explanations, prefixes, suffixes or reasoning content.

```json
{
  "Case Clarity": <1-5>,
  "Completeness of Case Elements": <1-5>,
  "Question-Answer Relevance": <1-5>,
  "Judgment Structural Compliance": <1-5>,
  "Legal Citation Format": <1-5>,
  "Answer Completeness": <1-5>,
  "Internal Consistency": <1-5>,
  "Substantiality of Answer": <1-5>,
  "Suggestion": "<Retain | Reject>",
  "Reason": "<Brief explanation within 50 words>"
}
```

\textbf{Judgment Rules}
- Retain: All scores $\ge 4$ and no score equals 1.
- Reject: Average score ranges from 3.0 to 3.9, or any item scores 2.
- Reject: Average score $< 3.0$, or any item scores 1.

\textbf{Question (Factual Findings)}
\_\_QUESTION\_\_

\textbf{Answer (Court's Opinions + Judgment Rulings)}
\_\_ANSWER\_\_

Please output the JSON scores strictly in the above format without any extra text outside the markdown code block:
\end{promptbox}

%% file: sections/prompts/rubric_pretag.tex

\begin{promptbox}{Rubric Checklist Annotation (Stage 1)}
You are a senior legal expert tasked with creating an **evaluation checklist** for a legal LLM hallucination benchmark.

[Your Task]
Given a legal question (question + manually labeled ground truth + metadata), extract **evaluation anchors** (per-subclass checklist) for each Layer-1 hallucination subclass based on the ground truth.

[Key Constraints]
1. The ground truth is authoritative. **Do not question, revise or add new conclusions**. You only organize existing content in the ground truth into structured check items.
2. Your output will serve as the rubric for downstream LLM-as-judge evaluation. Every item must be **verifiable by another judge**.
3. **Do not omit any Layer-1 subclass**. All 19 subclasses must be included (see schema below).
   Note: In version v0.3.2, L1.1 is reorganized into four consecutive subclasses (L1.1.1 \textasciitilde{} L1.1.4). Strictly follow the numbers and field names specified here; do not use old numbering.
4. If a subclass is irrelevant to the question, set `"applicable": false` and add a one-sentence `reason`. Do not remove the entire field.
5. Output a single JSON object only. No Markdown code blocks, no extra explanations, no ```json``` wrappers.
6. Use double quotes for all strings in JSON. No trailing commas. Keep all Chinese text unchanged.

[Overview of Layer-1 Subclass Schema]
All subclasses must be present, and each contains the field `"applicable"` (true/false).
- When `"applicable": true`, fill in all exclusive fields of the subclass as specified below; no omissions allowed.
- When `"applicable": false`, only fill in the `"reason"` field with one sentence explaining irrelevance.
- Fill in [] or null for any field that cannot be inferred from the question. Do not fabricate content.

[L1.1.1 Source Fabrication]
  schema:
    "must_cite_sources": [
      {"ref": "Civil Code Article 169" | "Criminal Law Article 201" | "Judicial Interpretation [2020] No.22" | "...",
       "core": "One-sentence summary of the key points of this provision or interpretation",
       "source_type": "statute|interpretation|guiding_case|regulation|meeting_minutes|textbook|answer_key"}
    ]
    (List all authoritative sources explicitly stated in the ground truth; use [] if none.
    **Do not add statutes, cases or interpretations absent from the ground truth**.)

[L1.1.2 Citation-Content Misapplication]
  schema:
    "expected_law_summaries": [
      {"ref": "Civil Code Article 169", "summary": "Core rule of this article (<=40 characters)"}
    ]
    "common_misapplications_to_avoid": ["..."]   # Similar provisions likely to be misquoted by the model (can be empty)

[L1.1.3 Hierarchy Error]
  schema:
    "hierarchy_constraints": ["Shall not apply Rule X against Rule Y", "This case applies departmental law / superior law / ..."]
    (Fill in only if the case involves conflicts of legal hierarchy; set applicable=false otherwise)

[L1.1.4 Granularity Error]
  schema:
    "exact_articles_required": [
      {"ref": "Civil Code Article 169 Paragraph 1", "exact_section": "Paragraph 1" | "Item (2)" | "Whole Article"}
    ]
    (List provisions specified down to paragraphs, items or sub-items in the ground truth; use [] for provisions only marked at article level)

[L1.2.1 Conceptual Confusion]
  schema:
    "core_concepts": ["Apparent Agency"]                     # Core legal concepts to be identified in this question
    "common_confusions_to_avoid": ["Unauthorized Agency", "Entrusted Agency"]   # Similar concepts easily confused by the model

[L1.2.2 Element Misstatement]
  schema:
    "must_have_elements": [
      {"element": "Counterparty acts in good faith", "meaning": "Explanation within 25 characters"}
    ]
    "elements_NOT_required": ["Counterparty is at fault"]   # Elements often wrongly added by the model

[L1.2.3 Exception Omission]
  schema:
    "must_mention_exceptions": ["Apparent agency is not established if the counterparty acts in bad faith", "..."]

[L1.2.4 Consequence Error]
  schema:
    "expected_consequences": ["The principal shall bear the legal effect of the agency act"]
    "common_wrong_consequences": ["Order to return property instead of awarding compensation"]

[L1.2.5 Doctrinal Position Confusion]
  schema:
    "required_position": "Prevailing Doctrine" | "Position stated in the Supreme People's Court Interpretation XX" | null
    "competing_positions_to_avoid": ["Minority Doctrine X"]

[L1.2.6 Discretionary Judgement Error]
  schema:
    "discretionary_anchors": [
      {"factor": "Circumstances are serious", "reasonable_range_or_anchor": "Description within 30 characters"}
    ]
    (Fill in only if the question obviously involves judicial discretion)

[L1.3.1 Jurisdiction Error]
  schema:
    "correct_jurisdiction": {
       "level": "Primary People's Court | Intermediate People's Court | Higher People's Court | Supreme People's Court | null",
       "region": "e.g. 'Court of the defendant's domicile' | 'Court of contract performance place'",
       "special_jurisdiction": "Explanation of exclusive jurisdiction | null"
    }

[L1.3.2 Period Error]
  schema:
    "required_periods": [
      {"name": "Limitation of action", "value": "3 years"},
      {"name": "Appeal period", "value": "15 days (civil cases)"}
    ]

[L1.3.3 Procedural-Step Error]
  schema:
    "required_steps": ["Mediation -> Case filing -> Evidence presentation -> Court hearing -> Judgment"]
    "common_step_mistakes_to_avoid": ["Preservation can only be applied after litigation (Incorrect, pre-litigation preservation is allowed)"]

[L1.3.4 Procedural-Outcome Error]
  schema:
    "expected_outcome": "Dismiss the case filing | Dismiss the claims | Voluntary withdrawal of lawsuit | Default judgment | null"
    "forbidden_outcomes": ["Dismiss the claims while the correct ruling is dismissing the case filing"]

[L1.3.5 Appeal Error]
  schema:
    "correct_appeal_path": "e.g. 'File an appeal to the Intermediate People's Court for second instance' | 'Apply for administrative reconsideration first before filing a lawsuit'"

[L1.4.1 Fact Fabrication]
  schema:
    "must_use_facts_from_question": ["Key Fact 1 from the question", "..."]
    "facts_NOT_in_question_must_avoid": ["Any amount, time, subject or conduct not stated in the question"]

[L1.4.2 Fact Omission]
  schema:
    "critical_facts_must_address": ["Key Fact 1 that determines the conclusion", "..."]
    (These are factual anchors relied on by the ground truth; omission will definitely lead to wrong answers)

[L1.4.3 Element-Fact Mismatch]
  schema:
    "element_fact_pairs": [
      {"element": "Counterparty acts in good faith", "evidence_in_question": "The third party had no knowledge of Xiao Zhang's ultra vires act"},
      {"element": "Appearance of authorization exists", "evidence_in_question": "The company's official seal was affixed"}
    ]

[L1.4.4 Party Confusion]
  schema:
    "parties_and_roles": [
      {"name": "Xiao Zhang", "role": "Unauthorized agent"},
      {"name": "Company", "role": "Principal"},
      {"name": "Third party", "role": "Counterparty"}
    ]

[Top-level Output Structure]
{
  "difficulty": "low | medium | high",
  "summary_anchors": "Core checkpoints summary (<=80 characters, remind judges of key evaluation points)",
  "L1_1_authority":      { L1_1_1_*, L1_1_2_*, L1_1_3_*, L1_1_4_* },   /* 4 subclasses in sequence */
  "L1_2_doctrine":       { L1_2_1_*, L1_2_2_*, L1_2_3_*, L1_2_4_*, L1_2_5_*, L1_2_6_* },
  "L1_3_procedural_law": { L1_3_1_*, L1_3_2_*, L1_3_3_*, L1_3_4_*, L1_3_5_* },
  "L1_4_application":    { L1_4_1_*, L1_4_2_*, L1_4_3_*, L1_4_4_* },
  "external_refs": [ {"ref": "...", "type": "statute|interpretation|..."} ]
}

[Few-shot Example]
INPUT:
[Question]
Xiao Zhang is a salesperson of a company. The company did not authorize him to sign contracts worth more than 500,000 yuan. Without obtaining authorization, Xiao Zhang signed a sales contract worth 1 million yuan with a client in the company's name and affixed the company's official seal on the contract. Before signing the contract, the client verified Xiao Zhang's identity with the company's finance department, which confirmed Xiao Zhang's position but did not specify the scope of authorization. Is this contract valid?

[ground_truth]
Answer: The contract is valid, and the company shall assume corresponding contractual obligations.
Legal Basis: Article 172 of the Civil Code of the People's Republic of China stipulates: "Where a person acts without power of agency, beyond the scope of power of agency, or after the termination of power of agency, and still performs an act of agency, if the counterparty has reason to believe that the person has power of agency, the act of agency shall be valid."
Analysis: This case constitutes apparent agency for the following reasons: (1) Xiao Zhang signed the contract in the company's name with the company's official seal, forming an apparent authorization; (2) The client verified Xiao Zhang's identity with the company's finance department before signing the contract, acting as a bona fide counterparty with reasonable care; (3) Transaction security based on the client's reasonable reliance shall be protected if the company denies the validity. Therefore, the contract is binding on the company, and the company may recover losses from Xiao Zhang after assuming liabilities.

[Metadata]
- Legal Field: Civil Law
- Task Form: Subjective Question
- Task Type: Legal Consultation

OUTPUT:
{
  "difficulty": "medium",
  "summary_anchors": "Core point: Determination of apparent agency. Analyze three elements including apparent authorization, bona fide counterparty and reasonable reliance per Civil Code Article 172; conclude the contract is valid.",
  "L1_1_authority": {
    "L1_1_1_source_fabrication": {
      "applicable": true,
      "must_cite_sources": [
        {"ref": "Civil Code Article 172", "core": "Rules for determination and legal effect of apparent agency", "source_type": "statute"}
      ]
    },
    "L1_1_2_citation_content_misapplication": {
      "applicable": true,
      "expected_law_summaries": [
        {"ref": "Civil Code Article 172", "summary": "Unauthorized agency is valid if the counterparty reasonably believes agency authority exists"}
      ],
      "common_misapplications_to_avoid": ["Civil Code Article 171 (ordinary unauthorized agency)"]
    },
    "L1_1_3_hierarchy_error": {
      "applicable": false,
      "reason": "This case only applies the Civil Code with no conflicts of legal hierarchy."
    },
    "L1_1_4_granularity_error": {
      "applicable": true,
      "exact_articles_required": [
        {"ref": "Civil Code Article 172", "exact_section": "Whole Article"}
      ]
    }
  },
  "L1_2_doctrine": {
    "L1_2_1_conceptual_confusion": {
      "applicable": true,
      "core_concepts": ["Apparent Agency"],
      "common_confusions_to_avoid": ["Ordinary Unauthorized Agency (Civil Code Article 171)", "Entrusted Agency", "Functional Agency"]
    },
    "L1_2_2_element_misstatement": {
      "applicable": true,
      "must_have_elements": [
        {"element": "Existence of apparent authorization", "meaning": "Objective appearance indicating valid agency authority"},
        {"element": "Counterparty acts in good faith", "meaning": "Counterparty is unaware of lack of agency authority"},
        {"element": "Reasonable reliance", "meaning": "Reliance formed after the counterparty exercises due care"}
      ],
      "elements_NOT_required": ["Counterparty is at fault", "Principal is at fault"]
    },
    "L1_2_3_exception_omission": {
      "applicable": true,
      "must_mention_exceptions": ["Apparent agency is not established if the counterparty acts in bad faith"]
    },
    "L1_2_4_consequence_error": {
      "applicable": true,
      "expected_consequences": ["The agency act binds the principal", "The principal may recover losses from the unauthorized agent after performance"],
      "common_wrong_consequences": ["The contract is invalid", "The contract is pending validity", "The unauthorized agent bears liabilities alone"]
    },
    "L1_2_5_doctrinal_position_confusion": {
      "applicable": false,
      "reason": "Apparent agency follows the prevailing doctrine with no competing theories to distinguish."
    },
    "L1_2_6_discretionary_judgement_error": {
      "applicable": true,
      "discretionary_anchors": [
        {"factor": "Reasonable reliance", "reasonable_range_or_anchor": "The client verified identity and fulfilled the duty of reasonable care"}
      ]
    }
  },
  "L1_3_procedural_law": {
    "L1_3_1_jurisdiction_error": {"applicable": false, "reason": "This is a substantive law consultation involving no jurisdiction issues."},
    "L1_3_2_period_error":       {"applicable": false, "reason": "This case involves no limitation periods or time limits."},
    "L1_3_3_procedural_step_error": {"applicable": false, "reason": "This case involves no litigation procedures."},
    "L1_3_4_procedural_outcome_error": {"applicable": false, "reason": "This case involves no procedural rulings."},
    "L1_3_5_appeal_error":       {"applicable": false, "reason": "This case involves no remedy or appeal procedures."}
  },
  "L1_4_application": {
    "L1_4_1_fact_fabrication": {
      "applicable": true,
      "must_use_facts_from_question": [
        "The company forbids Xiao Zhang from signing contracts exceeding 500,000 yuan",
        "Xiao Zhang signed a 1 million-yuan contract in the company's name",
        "The company's official seal was affixed on the contract",
        "The client verified Xiao Zhang's identity with the company's finance department",
        "The finance department confirmed Xiao Zhang's position but not the authorization scope"
      ],
      "facts_NOT_in_question_must_avoid": ["Any amount, time, subject or conduct not stated in the question"]
    },
    "L1_4_2_fact_omission": {
      "applicable": true,
      "critical_facts_must_address": [
        "The company's official seal was affixed (key evidence for apparent authorization)",
        "The client verified identity with the finance department (key evidence for reasonable reliance)"
      ]
    },
    "L1_4_3_element_fact_mismatch": {
      "applicable": true,
      "element_fact_pairs": [
        {"element": "Existence of apparent authorization", "evidence_in_question": "The company's official seal was affixed on the contract"},
        {"element": "Counterparty acts in good faith", "evidence_in_question": "The client verified identity and received no denial"},
        {"element": "Reasonable reliance", "evidence_in_question": "The client fulfilled the duty of inquiry and due care"}
      ]
    },
    "L1_4_4_party_confusion": {
      "applicable": true,
      "parties_and_roles": [
        {"name": "Xiao Zhang", "role": "Unauthorized agent / Company salesperson"},
        {"name": "Company", "role": "Principal"},
        {"name": "Client", "role": "Counterparty"}
      ]
    }
  },
  "external_refs": [
    {"ref": "Civil Code Article 172", "type": "statute"}
  ]
}

--- Per-item template tail ---

[Question]
<<QUESTION>>

[ground_truth]
<<GROUND_TRUTH>>

[Metadata]
- Legal Field: <<LEGAL_CATEGORY>>
- Task Form: <<TASK_FORM>>
- Task Type: <<LEGAL_TASK_TYPE>>

Output a pure JSON object covering all 4 top-level categories and 19 subclasses as required:
\end{promptbox}

%% file: sections/prompts/judge_L1_1_authority.tex
\begin{promptbox}{Judge Prompt: L1.1 Authority}
You are an expert in evaluating legal hallucinations in China.
- You only judge whether the evaluated rollout makes mistakes in this subclass.
- Judgments shall be based strictly on the ground truth (gt) and rubric specified in the checklist; do not add content beyond the rubric.
- If the rubric marks `applicable=false`, this subclass is not applicable. Directly set `hit=false` and leave the evidence blank.
- Do not add extra explanatory content; your only task is to identify errors.
- Output a single standard JSON object only, with no extra text outside markdown code blocks.

[Group] L1.1 Authority (Source/Authority Hallucination) (L1_1_authority)

[Subclass Guidance]
Key Judgment Points:
- Prioritize verification against must_cite_sources and expected_law_summaries listed in the rubric.
- L1.1.1 Source Fabrication: The model cites non-existent laws, judicial interpretations or case numbers outside the rubric list. Fill cited_source with the original text from the model.
- L1.1.2 Citation-Content Misapplication: The cited source is authentic but (1) inapplicable to the case (2) described content inconsistent with the original text (3) wrong article number.
- L1.1.3 Hierarchy Error: Applying lower-level laws against higher-level laws, or classifying administrative regulations as laws.
- L1.1.4 Granularity Error: Correct law name and article number, yet incorrect paragraph, item or sub-item.

Mutually Exclusive Classification (One error corresponds to only one category, follow priority below):
Non-existent source -> L1.1.1
Authentic source used incorrectly (inapplicable / inconsistent content / wrong article number) -> L1.1.2
Correct law name and article number, only wrong paragraph/item/sub-item -> L1.1.4
Note: Wrong article number (e.g. citing Article 35 instead of Article 36) belongs to L1.1.2 rather than L1.1.4.

Counterexamples (all marked hit=false):
- Incomplete coverage: The rollout cites part of must_cite_sources with fully correct content and conclusions consistent with ground truth.
- Wrong final answer but correct cited text: Do not mark L1.1 as hit merely because the final answer is incorrect.

[User prompt template]
## Group: <<GROUP>>
## Framework: <<FRAMEWORK>>    Dataset: <<DATASET>>
## Trajectory summary
<<TRAJECTORY_SUMMARY>>

## Question
<<QUESTION>>

## Ground truth (gt)
<<GROUND_TRUTH>>

## Checklist (rubric -- sliced to this group)
<<RUBRIC_JSON>>

## Rollout trajectory (rendered, step-marked)
<<TRAJECTORY_MD>>

## Rollout final answer
<<FINAL_ANSWER>>

## Subclass definitions
<<GUIDANCE>>

**Required output JSON schema**
{
  "L1_1_1_source_fabrication": {
    "hit": <true|false>,
    "step_indices": [<int>, ...],
    "origin_step": <int|null>,
    "evidence": "<Quote original content and state error in one sentence>",
    "cited_source": "<Original text of non-existent source; empty string if hit=false>"
  },
  "L1_1_2_citation_content_misapplication": {
    "hit": <true|false>,
    "step_indices": [<int>, ...],
    "origin_step": <int|null>,
    "evidence": "<Quote original content and state error in one sentence>",
    "misapplied_ref": "<Original text of misapplied source>"
  },
  "L1_1_3_hierarchy_error": {
    "hit": <true|false>,
    "step_indices": [<int>, ...],
    "origin_step": <int|null>,
    "evidence": "<Quote original content and state error in one sentence>",
    "hierarchy_issue": "<Specific description of hierarchy or validity error>"
  },
  "L1_1_4_granularity_error": {
    "hit": <true|false>,
    "step_indices": [<int>, ...],
    "origin_step": <int|null>,
    "evidence": "<Quote original content and state error in one sentence>",
    "granularity_detail": "<Specific description of wrong paragraph/item/sub-item>"
  }
}

Strictly return a single JSON object complying with the above schema. All subclasses must be retained.
\end{promptbox}

%% file: sections/prompts/judge_L1_2_doctrine.tex

\begin{promptbox}{Judge Prompt: L1.2 Doctrine}

You are an expert evaluator for legal hallucinations of Chinese laws (LLM-as-a-Judge).
- You only assess whether the evaluated rollout makes errors in corresponding subcategories.
- Judgments shall be strictly based on the ground truth (gt) and rubric in the checklist; do not add extra content beyond the rubric.
- If the rubric marks applicable=false, the subcategory is not applicable. Directly set hit=false and leave the evidence blank.
- Do not add redundant explanations; only identify existing errors.
- Output solely a standard JSON object, with no extra text outside the JSON.

[Group] L1.2 Doctrine (Doctrinal & Constituent Element Hallucination) (L1_2_doctrine)

[Subclass guidance]
Key Judgment Points:
- L1.2.1 Conceptual Confusion vs L1.2.2 Element Misstatement: First check if the model misidentifies the overall legal concept.
If the model confuses Concept A with Concept B and conducts the whole reasoning based on Concept B -> L1.2.1.
If the concept is correctly identified but individual constituent elements are misstated -> L1.2.2.
- L1.2.3 Exception Omission: Mark hit=true only when the exceptions listed in rubric.must_mention_exceptions directly determine the answer, the rollout completely ignores them, and the final answer is wrong as a result.
Counterexamples: The rollout lists partial exceptions while the answer is consistent with gt -> hit=false; equivalent expressions for exceptions are used -> hit=false.
- L1.2.4 Consequence Error: Wrong legal consequences (e.g., ruling for compensation instead of restitution, ruling revocable instead of invalid).
- L1.2.5 Doctrinal Position Confusion: Repeated shifts of legal positions within one single output.
- L1.2.6 Discretionary Judgement Error: Misjudgment on the boundary of indeterminate legal concepts (e.g., "seriously improper", "obviously unfair").

[User prompt template]
## Group: <<GROUP>>
## Framework: <<FRAMEWORK>>    Dataset: <<DATASET>>
## Trajectory summary
<<TRAJECTORY_SUMMARY>>

## Question
<<QUESTION>>

## Ground truth (gt)
<<GROUND_TRUTH>>

## Checklist (rubric -- sliced to this group)
<<RUBRIC_JSON>>

## Rollout trajectory (rendered, step-marked)
<<TRAJECTORY_MD>>

## Rollout final answer
<<FINAL_ANSWER>>

## Subclass definitions
<<GUIDANCE>>

Required output JSON schema
{
  "L1_2_1_conceptual_confusion": {
    "hit": <true|false>,
    "step_indices": [<int>, ...],
    "origin_step": <int|null>,
    "evidence": "<Quote original text and state the error in one sentence>",
    "confused_pair": "<Specify the mistaken substitution between Concept A and Concept B>"
  },
  "L1_2_2_element_misstatement": {
    "hit": <true|false>,
    "step_indices": [<int>, ...],
    "origin_step": <int|null>,
    "evidence": "<Quote original text and state the error in one sentence>",
    "wrong_element": "<Specify the misstated constituent element>"
  },
  "L1_2_3_exception_omission": {
    "hit": <true|false>,
    "step_indices": [<int>, ...],
    "origin_step": <int|null>,
    "evidence": "<Quote original text and state the error in one sentence>",
    "missing_exception": "<Specify the omitted exception or defense>"
  },
  "L1_2_4_consequence_error": {
    "hit": <true|false>,
    "step_indices": [<int>, ...],
    "origin_step": <int|null>,
    "evidence": "<Quote original text and state the error in one sentence>",
    "wrong_consequence": "<Specify the incorrect legal consequence or liability form>"
  },
  "L1_2_5_doctrinal_position_confusion": {
    "hit": <true|false>,
    "step_indices": [<int>, ...],
    "origin_step": <int|null>,
    "evidence": "<Quote original text and state the error in one sentence>",
    "doctrinal_issue": "<Specify where positions shift or conflicting theories are mixed>"
  },
  "L1_2_6_discretionary_judgement_error": {
    "hit": <true|false>,
    "step_indices": [<int>, ...],
    "origin_step": <int|null>,
    "evidence": "<Quote original text and state the error in one sentence>",
    "discretion_issue": "<Specify the misjudgment on discretionary boundaries>"
  }
}

Strictly return a single JSON object complying with the above schema. All subcategories must be retained.
\end{promptbox}

%% file: sections/prompts/judge_L1_3_procedural.tex

\begin{promptbox}{Judge Prompt: L1.3 Procedural-Law}
[System prompt]
You are an expert evaluator for legal hallucinations of Chinese laws (LLM-as-a-Judge).
- You only assess whether the evaluated rollout makes errors in corresponding subcategories.
- Judgments shall be strictly based on the ground truth (gt) and rubric in the checklist; do not add extra content beyond the rubric.
- If the rubric marks `applicable=false`, the subcategory is not applicable. Directly set `hit=false` and leave the evidence blank.
- Do not add redundant explanations; only identify existing errors.
- Output solely a standard JSON object, with no extra text outside the JSON.

[Group] L1.3 Procedural-Law (Procedural Hallucination) (L1\_3\_procedural)

[Subclass guidance]
Key Judgment Points:
- L1.3.1 Jurisdiction: Errors regarding tier jurisdiction, territorial jurisdiction, exclusive jurisdiction or agreement jurisdiction.
- L1.3.2 Period: Errors concerning time limits, including 3-year statute of limitations, 15-day appeal period, 3-month second-instance trial period and 6-month retrial application period. Judge directly by numerical accuracy.
- L1.3.3 Procedural-Step: Errors in procedural steps or sequence.
- L1.3.4 Procedural-Outcome: Common civil procedure errors in China. Distinguish "dismissal of suit" (procedural issue) and "dismissal of claims" (substantive issue) and mark separately.
- L1.3.5 Appeal: Errors in remedy approaches, such as administrative reconsideration prior to litigation and special remedies against arbitration awards.

[User prompt template]
## Group: <<GROUP>>
## Framework: <<FRAMEWORK>>    Dataset: <<DATASET>>
## Trajectory summary
<<TRAJECTORY_SUMMARY>>

## Question
<<QUESTION>>

## Ground truth (gt)
<<GROUND_TRUTH>>

## Checklist (rubric -- sliced to this group)
<<RUBRIC_JSON>>

## Rollout trajectory (rendered, step-marked)
<<TRAJECTORY_MD>>

## Rollout final answer
<<FINAL_ANSWER>>

## Subclass definitions
<<GUIDANCE>>

\textbf{Required output JSON schema}
\{
  "L1\_3\_1\_jurisdiction\_error": \{
    "hit": <true|false>,
    "step\_indices": [<int>, ...],
    "origin\_step": <int|null>,
    "evidence": "<Quote original text and state the error in one sentence>",
    "jurisdiction\_issue": "<Specify the specific jurisdiction error>"
  \},
  "L1\_3\_2\_period\_error": \{
    "hit": <true|false>,
    "step\_indices": [<int>, ...],
    "origin\_step": <int|null>,
    "evidence": "<Quote original text and state the error in one sentence>",
    "period\_issue": "<Specify the wrong time limit or limitation period>"
  \},
  "L1\_3\_3\_procedural\_step\_error": \{
    "hit": <true|false>,
    "step\_indices": [<int>, ...],
    "origin\_step": <int|null>,
    "evidence": "<Quote original text and state the error in one sentence>",
    "step\_issue": "<Specify the specific procedural step error>"
  \},
  "L1\_3\_4\_procedural\_outcome\_error": \{
    "hit": <true|false>,
    "step\_indices": [<int>, ...],
    "origin\_step": <int|null>,
    "evidence": "<Quote original text and state the error in one sentence>",
    "outcome\_issue": "<Specify the procedural conclusion error, e.g. dismissing claims instead of dismissing suit>"
  \},
  "L1\_3\_5\_appeal\_error": \{
    "hit": <true|false>,
    "step\_indices": [<int>, ...],
    "origin\_step": <int|null>,
    "evidence": "<Quote original text and state the error in one sentence>",
    "appeal\_issue": "<Specify the specific remedy approach error>"
  \}
\}

Strictly return a single JSON object complying with the above schema. All subcategories must be retained.
\end{promptbox}

%% file: sections/prompts/judge_L1_4_application.tex

\begin{promptbox}{Judge Prompt: L1.4 Application \& Subsumption}

You are an expert evaluator for legal hallucinations of Chinese laws (LLM-as-a-Judge).
- You only assess whether the evaluated rollout makes errors in corresponding subcategories.
- Judgments shall be strictly based on the ground truth (gt) and rubric in the checklist; do not add extra content beyond the rubric.
- If the rubric marks `applicable=false`, the subcategory is not applicable. Directly set `hit=false` and leave the evidence blank.
- Do not add redundant explanations; only identify existing errors.
- Output solely a standard JSON object, with no extra text outside the JSON.

[Group] L1.4 Application \& Subsumption (Fact Subsumption Hallucination) (L1\_4\_application)

[Subclass guidance]
Key Judgment Points:
- L1.4.1 Fact Fabrication: The model states facts, amounts, time, parties or conducts that do not appear in the original question.
- L1.4.2 Fact Omission: Key facts from the original question are omitted.
- L1.4.3 Element-Fact Mismatch: Legal elements are correctly stated, but wrongly matched with case facts. Distinguish from L1.2.2, which refers to incorrect legal elements themselves.
- L1.4.4 Party Confusion: Mix up Party A and Party B, or confuse agents with the principal parties.

[User prompt template]
## Group: <<GROUP>>
## Framework: <<FRAMEWORK>>    Dataset: <<DATASET>>
## Trajectory summary
<<TRAJECTORY_SUMMARY>>

## Question
<<QUESTION>>

## Ground truth (gt)
<<GROUND_TRUTH>>

## Checklist (rubric -- sliced to this group)
<<RUBRIC_JSON>>

## Rollout trajectory (rendered, step-marked)
<<TRAJECTORY_MD>>

## Rollout final answer
<<FINAL_ANSWER>>

## Subclass definitions
<<GUIDANCE>>

\textbf{Required output JSON schema}
\{
  "L1\_4\_1\_fact\_fabrication": \{
    "hit": <true|false>,
    "step\_indices": [<int>, ...],
    "origin\_step": <int|null>,
    "evidence": "<Quote original text and state the error in one sentence>",
    "fabricated\_fact": "<Specify the fabricated fact, amount, time, subject or conduct>"
  \},
  "L1\_4\_2\_fact\_omission": \{
    "hit": <true|false>,
    "step\_indices": [<int>, ...],
    "origin\_step": <int|null>,
    "evidence": "<Quote original text and state the error in one sentence>",
    "omitted\_fact": "<Specify the omitted or misplaced key fact>"
  \},
  "L1\_4\_3\_element\_fact\_mismatch": \{
    "hit": <true|false>,
    "step\_indices": [<int>, ...],
    "origin\_step": <int|null>,
    "evidence": "<Quote original text and state the error in one sentence>",
    "mismatch\_detail": "<Specify the details of mismatch between legal elements and facts>"
  \},
  "L1\_4\_4\_party\_confusion": \{
    "hit": <true|false>,
    "step\_indices": [<int>, ...],
    "origin\_step": <int|null>,
    "evidence": "<Quote original text and state the error in one sentence>",
    "party\_issue": "<Specify the details of confusion among parties>"
  \}
\}

Strictly return a single JSON object complying with the above schema. All subcategories must be retained.
\end{promptbox}

%% file: sections/prompts/judge_L2_1_planning_reasoning.tex

\begin{promptbox}{Judge Prompt: L2.1 Planning \& Reasoning}
[System prompt]
You are an expert evaluator for legal hallucinations of Chinese laws (LLM-as-a-Judge).
- You only assess whether the evaluated rollout makes errors in corresponding subcategories.
- Judgments shall be strictly based on the ground truth (gt) and rubric in the checklist; do not add extra content beyond the rubric.
- If the rubric marks `applicable=false`, the subcategory is not applicable. Directly set `hit=false` and leave the evidence blank.
- Do not add redundant explanations; only identify existing errors.
- Output solely a standard JSON object, with no extra text outside the JSON.

[Group] L2.1 Planning \& Reasoning (Planning & Reasoning Hallucination) (L2\_1\_planning\_reasoning)

[Subclass guidance]
Key Judgment Points:
- L2.1.1 Premature Closure: Mark `hit=true` only if both conditions below are satisfied:
  (a) No genuine examination of other options, counterexamples or opposing arguments after the step where the final answer is given;
  (b) The thinking in that step fails to analyze all candidate options equally and merely rationalizes the predetermined answer.
Counterexamples (all marked `hit=false`):
  - Standard writing template: Present conclusion first followed by comprehensive analysis for all options.
  - Hypothesize-then-verify: A tentative answer is put forward and verified via tool calls or subsequent steps.
  - Intermediate hypothesis in multi-step reasoning with continuous analysis in later steps.
- L2.1.2 Syllogism: Broken or reversed logical chain of major premise (legal provision) -> minor premise (case facts) -> conclusion.
- L2.1.3 Self-Contradiction: Conflicting conclusions or element selections within a single output or across steps; inconsistent legal citations in different steps.
- L2.1.4 Step Skip / Conflation: Drawing a direct conclusion without step-by-step reasoning; jumping from case facts to judgment without corresponding legal provisions.
- L2.1.5 Out-of-Context Quoting: Citing partial content of legal provisions while ignoring preconditions and restrictive clauses.

[User prompt template]
\#\# Group: <<GROUP>>
\#\# Framework: <<FRAMEWORK>>    Dataset: <<DATASET>>
\#\# Trajectory summary
<<TRAJECTORY\_SUMMARY>>

\#\# Question
<<QUESTION>>

\#\# Ground truth (gt)
<<GROUND\_TRUTH>>

\#\# Checklist (rubric - sliced to this group)
<<RUBRIC\_JSON>>

\#\# Rollout trajectory (rendered, step-marked)
<<TRAJECTORY\_MD>>

\#\# Rollout final answer
<<FINAL\_ANSWER>>

\#\# Subclass definitions
<<GUIDANCE>>

\textbf{Required output JSON schema}
\{
  "L2\_1\_1\_premature\_closure": \{
    "hit": <true|false>,
    "step\_indices": [<int>, ...],
    "origin\_step": <int|null>,
    "evidence": "<Quote original text and state the error in one sentence>",
    "closure\_step": "<Step number where the conclusion is finalized>"
  \},
  "L2\_1\_2\_syllogism\_error": \{
    "hit": <true|false>,
    "step\_indices": [<int>, ...],
    "origin\_step": <int|null>,
    "evidence": "<Quote original text and state the error in one sentence>",
    "syllogism\_issue": "<Specify the faulty part among major premise, minor premise and conclusion>"
  \},
  "L2\_1\_3\_self\_contradiction": \{
    "hit": <true|false>,
    "step\_indices": [<int>, ...],
    "origin\_step": <int|null>,
    "evidence": "<Quote original text and state the error in one sentence>",
    "contradiction\_pair": "<Two conflicting excerpts from the text>"
  \},
  "L2\_1\_4\_step\_skip\_or\_conflation": \{
    "hit": <true|false>,
    "step\_indices": [<int>, ...],
    "origin\_step": <int|null>,
    "evidence": "<Quote original text and state the error in one sentence>",
    "skip\_issue": "<Specify the skipped or merged key reasoning steps>"
  \},
  "L2\_1\_5\_out\_of\_context\_quoting": \{
    "hit": <true|false>,
    "step\_indices": [<int>, ...],
    "origin\_step": <int|null>,
    "evidence": "<Quote original text and state the error in one sentence>",
    "oo\_context\_quote": "<The quoted excerpt taken out of context>"
  \}
\}

Strictly return a single JSON object complying with the above schema. All subcategories must be retained.
\end{promptbox}

%% file: sections/prompts/judge_L2_2_memory.tex

\begin{promptbox}{Judge Prompt: L2.2 Memory}
You are an expert evaluator for legal hallucinations of Chinese laws (LLM-as-a-Judge).
- You only assess whether the evaluated rollout makes errors in corresponding subcategories.
- Judgments shall be strictly based on the ground truth (gt) and rubric in the checklist; do not add extra content beyond the rubric.
- If the rubric marks `applicable=false`, the subcategory is not applicable. Directly set `hit=false` and leave the evidence blank.
- Do not add redundant explanations; only identify existing errors.
- Output solely a standard JSON object, with no extra text outside the JSON.

[Group] L2.2 Memory (Memory Hallucination) (L2\_2\_memory)

[Subclass guidance]
Key Judgment Points: L2.2 only covers errors related to memory loss or context forgetting, not all wrong answers.
Trigger conditions (satisfy any one):
- Restating case facts in later steps inconsistently with the original question, including wrong amounts, names, time or actors.
- Answering a different question than the user raised in the final answer (task drift).
- Losing key qualifiers when paraphrasing the question in tool call queries.
- Contradictory restatement of previously confirmed facts in multi-step reasoning.

Counterexamples (all marked `hit=false`):
- Wrong judgment on legal compliance of options -> Classified as L1.4 or L1.2.
- Misstating constituent elements of legal provisions -> Classified as L1.2.2.
- Incorrect final answer with accurate restatement of facts -> Reasoning or judgment error rather than memory error.
- Single thinking plus final answer structure (max_step <= 2 and n_events <= 3) generally does not trigger L2.2, unless obvious factual data conflicts with the original question.

[User prompt template]
## Group: <<GROUP>>
## Framework: <<FRAMEWORK>>    Dataset: <<DATASET>>
## Trajectory summary
<<TRAJECTORY_SUMMARY>>

## Question
<<QUESTION>>

## Ground truth (gt)
<<GROUND_TRUTH>>

## Checklist (rubric -- sliced to this group)
<<RUBRIC_JSON>>

## Rollout trajectory (rendered, step-marked)
<<TRAJECTORY_MD>>

## Rollout final answer
<<FINAL_ANSWER>>

## Subclass definitions
<<GUIDANCE>>

\textbf{Required output JSON schema}
\{
  "L2\_2\_memory": \{
    "hit": <true|false>,
    "step\_indices": [<int>, ...],
    "origin\_step": <int|null>,
    "evidence": "<Quote original text and state the error in one sentence>",
    "forgotten\_premise": "<Forgotten or misrecorded premise, name, amount or time>"
  \}
\}

Strictly return a single JSON object complying with the above schema. All subcategories must be retained.
\end{promptbox}

%% file: sections/prompts/judge_L2_3_tool_observation.tex

\begin{promptbox}{Judge Prompt: L2.3 Tool \& Observation}
You are an expert evaluator for legal hallucinations of Chinese laws (LLM-as-a-Judge).
- You only assess whether the evaluated rollout makes errors in corresponding subcategories.
- Judgments shall be strictly based on the ground truth (gt) and rubric in the checklist; do not add extra content beyond the rubric.
- If the rubric marks `applicable=false`, the subcategory is not applicable. Directly set `hit=false` and leave the evidence blank.
- Do not add redundant explanations; only identify existing errors.
- Output solely a standard JSON object, with no extra text outside the JSON.

[Group] L2.3 Tool \& Observation (Tool & Observation Hallucination) (L2\_3\_tool\_observation)

[Subclass guidance]
Judgments must comply with the following framework tool specifications without arbitrary interpretation.
- L2.3.1 Tool-Call Error:
  - Wrong tool selection: Violating tool selection rules, e.g. using law\_retrieval instead of law\_check with known article numbers, or retrying law\_retrieval when web\_search should be adopted.
  - Invalid parameters: Irrelevant query, incorrect parameter type or value, wrong law name.
  - Wrong calling order: Invoking tools exclusive to deep-analysis phase in earlier stages (only for lawthinker).
  - Excessive or missing calls: Repeated calls with identical tool and parameters, or failing to switch tools/terminate after >=3 unsuccessful attempts with the same query.
  - `MALFORMED` tagged tool call indicates JSON parsing failure, which shall be marked as hit=true.
- L2.3.2 Observation Misuse:
  - Only adopting the first retrieved legal provision out of multiple returned items.
  - Misreading figures such as confusing the year 2018 with 2008 in case records.
  - Drawing conclusions based on full text while only reading abstracts.

Counterexamples (all marked hit=false):
- Irrelevant provisions returned by law\_retrieval are inherent noise. Do not mark error unless irrelevant provisions are used for reasoning.
- This group is only applicable when there is at least one tool call event.

[User prompt template]
\#\# Group: <<GROUP>>
\#\# Framework: <<FRAMEWORK>>    Dataset: <<DATASET>>
\#\# Trajectory summary
<<TRAJECTORY\_SUMMARY>>

\#\# Question
<<QUESTION>>

\#\# Ground truth (gt)
<<GROUND\_TRUTH>>

\#\# Checklist (rubric - sliced to this group)
<<RUBRIC\_JSON>>

\#\# Rollout trajectory (rendered, step-marked)
<<TRAJECTORY\_MD>>

\#\# Rollout final answer
<<FINAL\_ANSWER>>

\#\# Subclass definitions
<<GUIDANCE>>

\#\# Framework tool spec (verbatim excerpt)
\#\#\# Tool List
- law\_retrieval: Legal provision retrieval. Returns top-k most relevant provisions given a natural language query. Parameters: \{"query": "str", "topk": "int"\}
  Usage rules:
  (1) Priority: Use law\_check for exact article numbers; use law\_retrieval only for natural language descriptions.
  (2) Query writing: Focus on crime/legal system names and key elements instead of full questions. topk is recommended 3-5.
  (3) Retry rule: Rewrite keywords once for poor results. Switch to web\_search or make conclusion directly if still failed. Repeated retries with identical query are forbidden.
  (4) Coverage: Current effective statutes, judicial interpretations and departmental rules of Chinese mainland. Low coverage for local regulations, newly released policies and foreign laws.
- law\_recommendation: Similar provision recommendation. Returns related provisions given a legal citation. Parameters: \{"law": "str"\}
- charge\_expansion: Related charge expansion. Returns similar charges given a list of charges. Parameters: \{"charges": "List[str]"\}
- case\_retrieval: Similar case retrieval. Returns analogous cases given case type and information. Parameters: \{"type": "str(Civil Case|Criminal Case)", "query": "str"\}
- template\_retrieval: Document template retrieval. Gets templates for specified legal documents. Parameters: \{"template\_type": "str"\}
- plan\_generation: Writing plan generation. Creates outlines for designated documents. Parameters: \{"document\_type": "str"\}
- procedure\_retrieval: Court procedure retrieval. Only for moot court scenarios. Parameters: \{"court\_type": "str(Civil Court|Criminal Court)", "stage": "int(0-4|0-2)"\}
- law\_check: Legal provision verification. Returns full text given exact legal citation. Parameters: \{"law\_name": "str"\}
  Usage rule: Prioritize law\_check when accurate article numbers are available for higher precision.
- fact\_law\_relevance\_check: Fact-provision relevance verification. Checks applicability of a given provision to case facts. Parameters: \{"fact": "str", "law": "str"\}
- crime\_law\_consistency\_check: Charge-provision matching verification. Validates correspondence between charges and criminal law articles. Parameters: \{"crime": "str", "law": "str"\}
- document\_format\_check: Document format inspection. Checks format of complete legal documents. Parameters: \{"document\_type": "str", "document": "str"\}
- law\_query\_rewrite: Query rewriting. Optimizes retrieval queries combining case background. Parameters: \{"query": "str", "context": "str"\}
- procedure\_check: Procedure inspection. Verifies completeness of court proceedings. Parameters: \{"court\_type": "str"\}
- web\_search: Web search. Retrieves legal information from public network.
  Application scenarios:
  (1) No valid results after one keyword rewrite with law\_retrieval / law\_check;
  (2) Inquiring about latest policies, local regulations, case details or foreign laws;
  (3) Consulting practical procedures, required materials and service channels.
  Discipline: Consolidate results and give final answer once relevant summaries are obtained. Repeated calls with identical query are prohibited.

\textbf{Required output JSON schema}
\{
  "L2\_3\_1\_tool\_call\_error": \{
    "hit": <true|false>,
    "step\_indices": [<int>, ...],
    "origin\_step": <int|null>,
    "evidence": "<Quote original text and state the error in one sentence>",
    "tool\_issue": "<Specify error type: wrong tool / invalid parameter / excessive calls etc.>"
  \},
  "L2\_3\_2\_observation\_misuse": \{
    "hit": <true|false>,
    "step\_indices": [<int>, ...],
    "origin\_step": <int|null>,
    "evidence": "<Quote original text and state the error in one sentence>",
    "misuse\_step": "<Step number where misinterpretation occurs>"
  \}
\}

Strictly return a single JSON object complying with the above schema. All subcategories must be retained.
\end{promptbox}